\documentclass[10pt,twocolumn,letterpaper]{article}

\usepackage{cvpr}
\usepackage{times}
\usepackage{epsfig}
\usepackage{graphicx}
\usepackage{amsmath}
\usepackage{amssymb}

\usepackage{multirow}
\usepackage{array}
\usepackage{dsfont}
\usepackage{boldline}

\usepackage{dblfloatfix}
\usepackage[titletoc,title]{appendix}

\usepackage[pagebackref=true,breaklinks=true,letterpaper=true,colorlinks,bookmarks=false]{hyperref}

\cvprfinalcopy 
\def\cvprPaperID{230}

\newcommand{\bst}[1]{\textcolor{red}{ \textbf{#1} }}
\newcommand{\second}[1]{\textcolor{blue}{ \underline{#1} }}

\ifcvprfinal\fi
\begin{document}

\title{Recalling Holistic Information for Semantic Segmentation}


\author{Hexiang Hu\thanks{equal contribution.}\\
UCLA\\
Los Angeles, CA\\
{\tt\small hexiang.frank.hu@gmail.com}
\and
Zhiwei Deng\footnotemark[1]\\
Simon Fraser University\\
Burnaby, BC, Canada\\
{\tt\small zhiweid@sfu.ca}
\and
Guang-Tong Zhou\\
Oracle Labs\\
Vancouver, BC, Canada\\
{\tt\small zhouguangtong@gmail.com}
\and
Fei Sha\\
UCLA \\
Los Angeles, CA\\
{\tt\small feisha@cs.ucla.edu}
\and
Greg Mori\\
Simon Fraser University\\
Burnaby, BC, Canada\\
{\tt\small mori@cs.sfu.ca}
}

\maketitle

\begin{abstract}
Semantic segmentation requires a detailed labeling of image pixels by object category. Information derived from local image patches is necessary to describe the detailed shape of individual objects. However, this information is ambiguous and can result in noisy labels. Global inference of image content can instead capture the general semantic concepts present. We advocate that high-recall holistic inference of image concepts provides valuable information for detailed pixel labeling. We build a two-stream neural network architecture that facilitates information flow from holistic information to local pixels, while keeping common image features shared among the low-level layers of both the holistic analysis and segmentation branches. We empirically evaluate our network on four standard semantic segmentation datasets. Our network obtains state-of-the-art performance on PASCAL-Context and NYUDv2, and ablation studies verify its effectiveness on ADE20K and SIFT-Flow.
\end{abstract}

\section{Introduction}
\label{sec:intro}

Image analysis is a fundamental problem in computer vision.  The task can be framed at different levels of granularity. At a fine scale, semantic segmentation labels each pixel to depict semantic elements by detailed shapes and contours.

\begin{figure}[t]
\begin{center}
\includegraphics[width=0.475\textwidth]{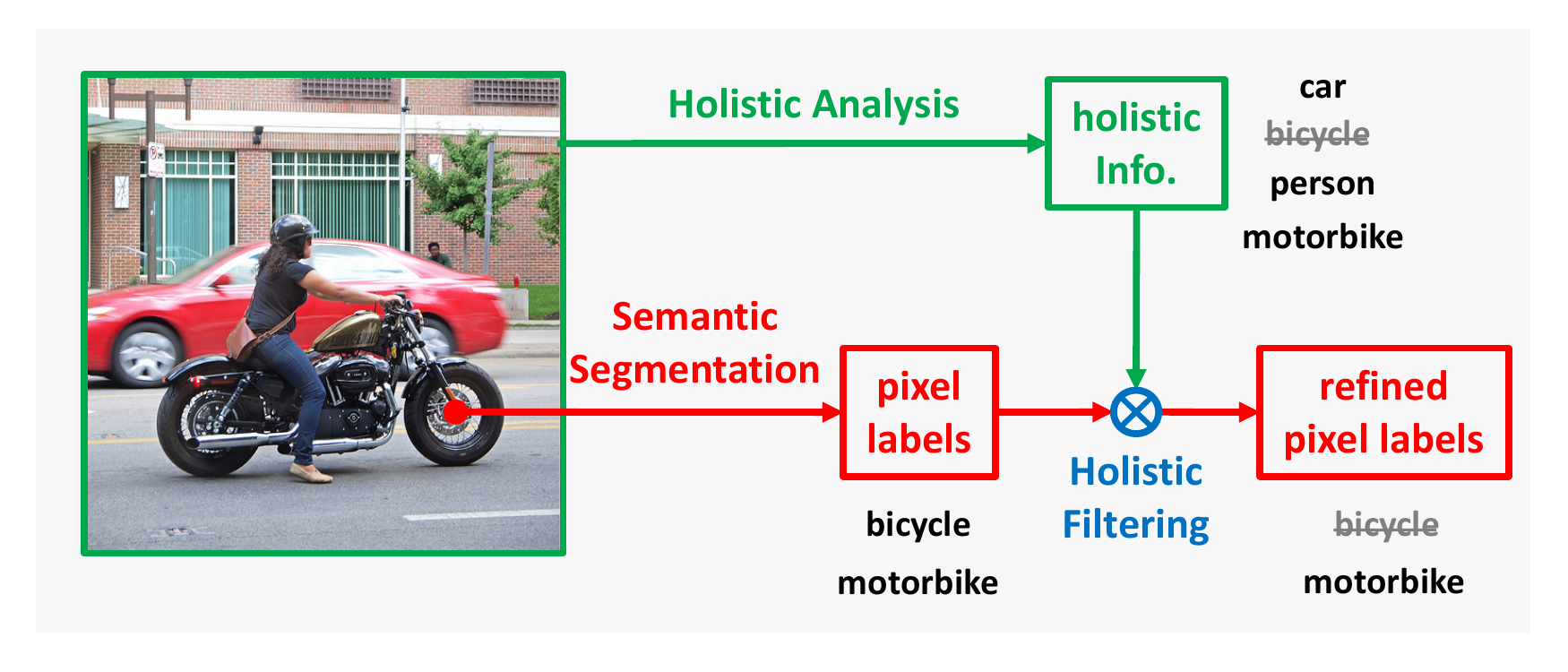}
\end{center}
\caption{An example showing the usage of holistic information in semantic segmentation. We obtain holistic information about the contents of the entire image, and leverage this to filter out noisy pixel predictions to improve semantic segmentation. In our example, the noisy pixel label (\textit{i.e.}, bicycle) is removed after holistic filtering since it is not recalled as a likely label via the holistic analysis.}
\label{fig:intro}
\vspace{-0.1in}
\end{figure}

However, detailed semantic segmentation is challenging -- there exists significant ambiguity in fine-scale image patches that can result in noisy semantic segmentation outputs.  The main focus of this paper is utilizing holistic information derived from analyzing entire images to filter noisy low-level semantic segmentation.
Holistic information about the content of an image is highly valuable for semantic segmentation on local pixels -- essentially, the existence, non-existence and co-existence information on semantic classes provides a strong cue in determining the local pixel labels (see Figure~\ref{fig:intro}). 

State-of-the-art methods for semantic segmentation leverage the successes of Convolutional Neural Networks (CNNs)~\cite{lecun98_pieee}. CNNs have transformed the field of image classification, especially since the development of AlexNet~\cite{alex_nips12}. There have been many follow-up CNN architectures to further boost image classification, including VGGNet~\cite{simonyan_iclr15}, Google Inception~\cite{szegedy_cvpr15}, ResNet~\cite{he_cvpr16}, \textit{etc}. Semantic segmentation utilizes these network structures, combined with dense output structures to label image pixels by semantic categories. A representative work is the Fully Convolutional Network (FCN)~\cite{shelhamer_pami16} that leverages skip features of CNNs to produce a detailed pixel labeling. Another example is the DeepLab~\cite{chen_arxiv16} framework, which augments FCN with dilated convolution~\cite{yu_iclr16}, atrous spatial pyramid pooling and Conditional Random Fields (CRFs), and obtains state-of-art semantic segmentation performance.




Common among these previous methods is a focus on (layers of) low-level pixel analysis leading to semantic segmentation.  State-of-the-art techniques combine this with sophisticated graphical model-style techniques (CRF) and pooling structures to obtain high accuracy.  The role of these additional components is to smooth out the noisy pixel labelings that result from the direct CNN analysis.  However, we believe that a simpler approach to incorporate this information is via a holistic analysis that globally suggests category labels that are likely to be present in the image.

To make use of holistic information in semantic segmentation, we leverage it to filter out noisy pixel predictions -- if the holistic information suggests a semantic category is unlikely to be present, then pixels should be unlikely to be predicted as that label. Of course, holistic image analysis is imperfect. We conduct a detailed problem analysis in Section~\ref{sec:analysis} to study how and when the holistic information can help.

We utilize these observations to propose a novel neural network architecture that unifies a holistic branch and a detailed semantic segmentation branch. The two branches share common image features in low-level layers of convolution and pooling. In the holistic branch, we then recall information about semantic categories by aggregating over a grid of image patches. In the segmentation branch, we pipe in holistic information to guide segmentation by filtering out noisy pixel predictions. Our network is end-to-end trainable towards the goal of high-quality semantic segmentation.

\noindent{\bf Contribution.} We summarize our main contributions as:
\vspace{-0.05in}
\begin{itemize}
\setlength{\itemsep}{0pt}
\setlength{\parskip}{0pt}
\setlength{\parsep}{0pt}
\item First, we advocate to leverage holistic image analysis to guide semantic segmentation, and provide empirical analysis to support our intuition.
\item Second, we invent \textit{holistic filtering} that enables us to filter out noisy pixel predictions with the guidance of holistic information.  
\item Third, we propose a two-stream neural network architecture for semantic segmentation. We implement a holistic branch and a segmentation branch, which facilitate the flow of global image information to the segmentation branch.
Our approach is general, and could be incorporated into a variety of CNN-based semantic segmentation architectures.
\item Finally, we evaluate our proposed network on PASCAL-Context~\cite{mottaghi_cvpr14}, ADE20K~\cite{zhou_arxiv16}, NYUDv2~\cite{silberman_eccv12} and SIFT-Flow~\cite{liu_pami11}. Experimental results show that the proposed network improves upon state-of-the-art semantic segmentation models such as FCN~\cite{shelhamer_pami16} and DeepLab~\cite{chen_arxiv16}.
\end{itemize}

\section{Related Work}
\label{sec:related}

\textbf{Convolutional neural networks}. In the last several years, CNNs have reformed the field of object recognition. The proposal of AlexNet~\cite{alex_nips12}, which achieves remarkable performance in ImageNet classification~\cite{russakovsky_ijcv15} with 8 layers of network, breaks the saturation bottleneck. Later on, Simonyan and Zisserman built VGGNet~\cite{simonyan_iclr15}, a deeper network architecture of 19 layers. Recently, He~\textit{et al.}~\cite{he_cvpr16} proposed ResNet with over a hundred layers by introducing residual connections for even better classification accuracy and more generalized image features.

\textbf{Semantic segmentation}. The significant success of CNNs in object recognition has led the recent new attention on semantic segmentation. A representative work is the Fully Convolutional Network (FCN)~\cite{shelhamer_pami16} that uses skip features of CNNs for detailed pixel labeling. FCN combines multi-level feature descriptors to leverage coarse-to-fine local pixel information. In a recent advance, the atrous convolution is introduced by Chen~\textit{et al.}~\cite{chen_arxiv16} as a technique to retain a large field of view while keeping fewer trainable weights in semantic segmentation networks. The same method termed as dilated convolution was also pursued by Yu and Koltun~\cite{yu_iclr16} by a cascading series of dilated convolution layers.

Another line of work pushes on refining the detailed shapes and contours of semantic segmentation. A fully connected CRF is proposed by Kr{\"{a}}henb{\"{u}}hl and Koltun~\cite{krahenbuhl_nips12} as an efficient dense pixel modeling method. The fully connected pairwise CRF is further adopted by Chen~\textit{et al.}~\cite{chen_arxiv16} and Zheng~\textit{et al.}~\cite{zheng_iccv15} on top of FCNs as a further refinement. These methods have achieved considerable improvement on the semantic image segmentation task. Different from the above-mentioned methods, our work leverages image-level holistic information to guide semantic segmentation, and we verify that correctly recalling holistic information is a key to improve the semantic segmentation performance.

\textbf{Global-local information fusion}. It has been shown that visual understanding benefits from exploiting and leveraging information of varying granularity. Deng~\textit{et al.}~\cite{deng_eccv14} modeled hierarchical and exclusive relations among semantic categories. Jain~\textit{et al.}~\cite{jain_cvpr16} developed recurrent neural network structures for spatio-temporal inference. Hu~\textit{et al.}~\cite{hu_cvpr16} proposed a neural graph inference model to propagate information among multiple levels of visual classes, including coarse labels, fine-grained categories as well as attributes. Amer~\textit{et al.}~\cite{amer_eccv12} adopted the and-or graph structure to reason about human activities at multiple levels of granularity. Gkioxari~\textit{et al.}~\cite{gkioxari_iccv15} exploited contextual cues to improve action recognition.

In the realm of semantic segmentation, He~\textit{et al.}~\cite{he_cvpr04} proposed to use multi-scale CRFs to capture features at various image resolutions for semantic segmentation. Approaches for modeling object instances and their segmentation have also been developed~\cite{kumar_cvpr05,wu_cvpr07}. Another example of this line improves instance-aware semantic segmentation with detection and classification~\cite{dai_cvpr15}. In contrast, we propose a framework to exploit holistic analysis for semantic segmentation.

\section{Problem Analysis}
\label{sec:analysis}

In this section, we would like to understand the effects of holistic information on semantic segmentation. We first design experiments to answer the question \textit{Can holistic information help semantic segmentation?} With a positive answer, we proceed to discover \textit{What makes holistic information helpful?} Given the observations, we also provide practical guidance for our network design.

\subsection{Can Holistic Information Help?}
\label{sec:analysis:holistic}


First, we conduct an analysis to determine whether holistic information can help in semantic segmentation. Suppose we knew the ground-truth image-level labels as the holistic information, how much performance gain would there be in semantic segmentation?  Addressing this question analyzes how much potential there is for incorporating holistic image-level reasoning into semantic segmentation.

We simply represent holistic information by the ground-truth labels assigned to each image, and then use the labels to filter out noisy segmentation outputs. Specifically, when labeling a pixel, we constrain the candidate set of labels to be the set of labels suggested by holistic information, \textit{i.e.}, the ground-truth image labels. As a comparison, the baseline methods uses all possible labels appearing in the dataset as the candidate labels. This simple change provides evidence of how much a holistic filtering could potentially help in semantic segmentation.

We apply existing semantic segmentation methods as baselines on two datasets. Specifically, we run FCN (8s) on PASCAL-Context~\cite{shelhamer_pami16} and DilatedNet on ADE20K~\cite{zhou_arxiv16}. We compare the performance of the baselines and their holistic filtering counterparts in Table~\ref{tab:analysis:holistic}. Please refer to Section~\ref{sec:experiment} for dataset details and the performance metrics.

\begin{table}[tp]
\small
\tabcolsep 3pt
\centering
\begin{tabular}{ c|cccc }
\multicolumn{5}{c}{(a). PASCAL-Context} \\
\hline
 & pAcc & mAcc & mIU & fwIU \\
\hline
FCN~\cite{shelhamer_pami16}  & 67.45 & 52.31 & 39.12 & 53.03 \\ 
FCN~\cite{shelhamer_pami16} + Holistic Filter & \bst{77.78} & \bst{62.42} & \bst{52.17} & \bst{65.01} \\
\hline
\multicolumn{5}{c}{} \\
\multicolumn{5}{c}{(b). ADE20K} \\
\hline
 & pAcc & mAcc & mIU & fwIU \\
\hline
DilatedNet~\cite{zhou_arxiv16} & 73.55 & 44.59 & 32.31 & 60.14 \\
DilatedNet~\cite{zhou_arxiv16} + Holistic Filter & \bst{81.22} & \bst{54.18} & \bst{45.19} & \bst{69.19} \\
\hline
\multicolumn{5}{c}{ }
\end{tabular}
\caption{Performance comparison of the baseline semantic segmentation methods and their holistic filtering counterparts. Here we use the ground-truth image labels as the holistic information. All measures are reported in percentage.}
\label{tab:analysis:holistic}
\end{table}

Table~\ref{tab:analysis:holistic} shows that the holistic filtering achieves favorable performance gain over baselines that do not use holistic information -- approximately 33\% and 40\% relative improvement on PASCAL-Context and ADE20K respectively in terms of mean IU. This is reasonable since the holistic information indicates what to expect in the image, and help to filter out noisy prediction on pixel labels.

Beyond leveraging holistic information in the post-segmentation process to filter out noisy pixel predictions, we have also retrained the network weights with the guidance of holistic information. It turns out that retraining brings additional performance gain. We will describe the detailed network design that enables retraining in Section~\ref{sec:method}, and further analyze our experimental results in Section~\ref{sec:experiment:analysis}.

\subsection{What Makes Holistic Information Helpful?}
\label{sec:analysis:recall}

We now continue to explore the key properties the holistic information should possess in order to help semantic segmentation. This exploration is valuable for practical usage because in a real scenario, we could never assume perfect image labels as the holistic information.

\begin{figure}[tp]
\begin{center}
\includegraphics[width=0.475\textwidth]{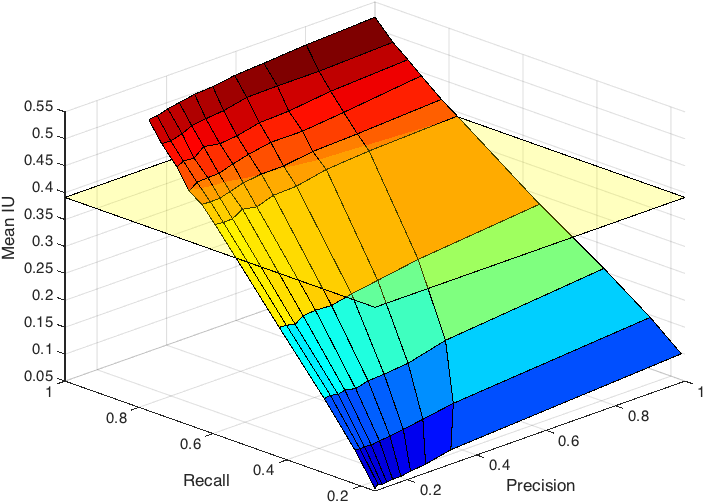}
\small{(a). PASCAL-Context}
\includegraphics[width=0.475\textwidth]{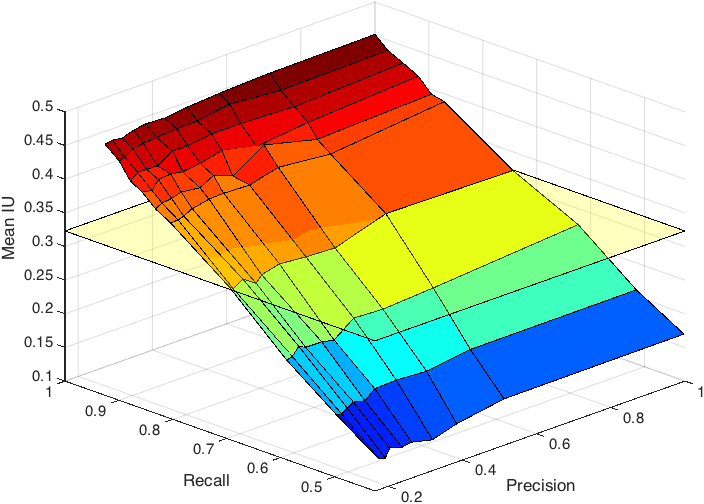}
\small{(b). ADE20K}
\end{center}
\caption{The mean IU grids of semantic segmentation with holistic filtering on the PASCAL-Context and ADE20K datasets, with respect to various precisions and recalls. The light yellow plane depicts the baseline performance, \textit{i.e.}, 39.12\% of FCN on PASCAL-Context~\cite{shelhamer_pami16} and 32.31\% of DilatedNet on ADE20K~\cite{zhou_arxiv16}. }
\label{fig:analysis:grid}
\end{figure}

We first design a mechanism to contaminate the holistic information. The idea is to start with the set of ground-truth image labels, and then randomly remove some ground-truth labels or add in noisy labels that are not in the ground-truth set. By performing such operations, we could effectively control the precision and recall of the ground-truth labels in the holistic information -- the precision is degraded when adding in more and more noisy labels, and the recall is degraded when removing more and more ground-truth labels.

\begin{figure*}[tp]
\centering
\includegraphics[width=1\textwidth]{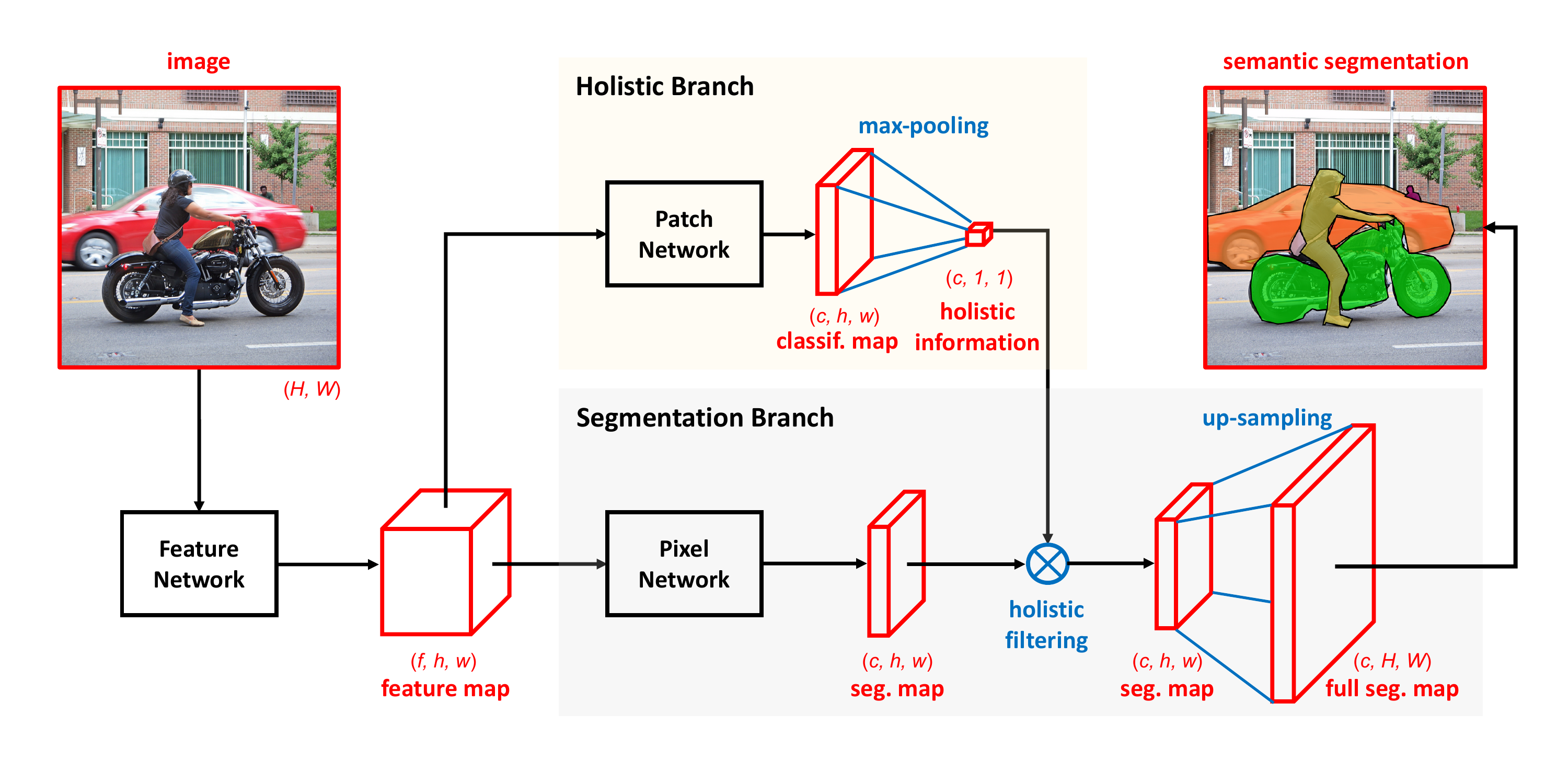}
\caption{Our network architecture for semantic segmentation. Please refer to the text for detailed description of the pipeline. The notations in brackets depict the size of the data. The input image size is $H \times W$. After the feature network, the image is down-sampled to $h \times w$ pixels. Moreover, we use $f$ to denote the number of feature channels, and $c$ to denote the number of pixel/image classes.}
\label{fig:method:arch}
\end{figure*}

We have experimented with the holistic filtering methods mentioned in Section~\ref{sec:analysis:holistic} with various precision and recall settings of ground-truth labels in holistic information (details of the experimental setup are provided in Appendix~\ref{sec:sup:pr}). We plot the mean IU performance on PASCAL-Context and ADE20K in Figure~\ref{fig:analysis:grid}.

The plots clearly show that the precision (of ground-truth labels in holistic information) has minor impact on the semantic segmentation performance -- the mean IU changes slightly as the precision degrades from 1 to 0.2. On the other hand, however, the recall (of ground-truth labels in holistic information) plays a key role for semantic segmentation -- the mean IU decreases significantly with respect to degradation of recall. Moreover, the comparison with the baseline methods shows that our holistic filtering method is able to outperform the baselines as long as the recall goes above 0.85 on PASCAL-Context and 0.75 on ADE20K. Here the holistic information is used only in the post segmentation process to filter out noisy pixel predictions -- the performance of our holistic filtering method is further boosted when retraining the network weights with the guidance of holistic information.

The above observations lead us to some practical guidelines for designing our semantic segmentation network. Specifically, we should have a holistic branch that recalls ground-truth image labels as much as possible to serve as the holistic information. If necessary, we would rather sacrifice precision in the predicted labels to improve recall. Furthermore, we need a segmentation branch that pipes the holistic information to filter semantic segmentation. The process needs to be differentiable, so that we can jointly train the holistic and segmentation branches end-to-end. Next we describe the details of our network design.

\section{Network Design}
\label{sec:method}

We now present our network design that leverages the guidelines derived from the previous problem analysis. Figure~\ref{fig:method:arch} illustrates the general network architecture.

In detail, we have a holistic branch to recall holistic information, and a segmentation branch to adopt holistic information to guide the semantic segmentation process. The two branches share a common feature network of convolutional layers and pooling layers. The feature network outputs a feature map of certain feature channels on down-sampled images. The feature network can be easily adopted from any existing CNN architecture (like VGGNet~\cite{simonyan_iclr15}, ResNet~\cite{he_cvpr16}, \textit{etc}.), by removing its top fully-connected layers and keeping the convolutional and pooling layers. This flexibility enables our network to take advantage of the continual improvement in CNN architectures. In our experiments, we use the 152-layer ResNet~\cite{he_cvpr16} by default, if not further specified.

In the following we describe our holistic branch and segmentation branch, respectively. After that, we will briefly go through our training process.

\subsection{Recall-Preserving Holistic Analysis}
\label{sec:method:classification}

Our holistic branch takes in the feature map as input and suggests image labels, with a focus on recalling ground-truth image labels as much as possible. To preserve high recall, we believe it is critical to conduct location-aware predictions that densely inspect image patches, since a semantic component (\textit{i.e.}, an object or stuff) always takes up a portion of the entire image. We then aggregate the location-aware predictions by max-pooling to compose the final holistic information.

Our holistic analysis is unlikely to miss ground-truth image labels via the dense location-aware prediction, and thus preserves high recall. However, it might incur noisy image labels since noisy or background image patches could mislead the max-pooling process. Note that our holistic analysis is different from standard image classification that optimizes both recall and precision, but it serves our purpose as we would rather sacrifice precision to improve recall.

In detail, we add in a patch network on top of the feature map to generate a classification map that predicts location-aware image labels using densely sliding image patches. The patch network is composed of three layers. First, we have a dilated convolutional layer~\cite{yu_iclr16,chen_arxiv16} to summarize image patch information. We use $k \times k$ convolutional kernels with a dilation rate of $r$ under $d$ channels. In our experiment, we set the default patch size as 224, $k=3$, $r=2$ and $d=512$, if not further specified. The second layer is a ReLU layer included for non-linearity. Finally, we top up with a one-by-one convolutional layer to predict a classification map of $c$ channels, each represents a semantic class.


We then compose the holistic information by a max-pooling over the classification map. The max-pooling results in a $c$-dimensional vector that describes the confidence scores for each semantic class to appear in the given image. These are then piped into the segmentation branch to guide semantic segmentation. Here we use soft label confidences to enable gradient back-propagation for end-to-end training.

\subsection{Semantic Segmentation with Holistic Filtering}
\label{sec:method:segmentation}

Our segmentation branch takes in the feature map and the holistic information to generate a semantic segmentation of the given image. As shown in Figure~\ref{fig:method:arch}, the branch is composed of three major processes including a pixel network, a holistic filtering and an up-sampling.

The pixel network is responsible for generating a segmentation map based on the feature map obtained from the feature network. The network structure can be easily adopted from any state-of-the-art semantic segmentation framework. In our experiments, we have evaluated two variants, FCN~\cite{shelhamer_pami16} and DilatedNet~\cite{yu_iclr16,chen_arxiv16} with slight modifications (details in Appendix~\ref{sec:sup:plus}). The FCN variant uses decovolution to generate three segmentation maps of various down-sampling rates. The DilatedNet applies dilated convolution to obtain a single segmentation map.

The holistic filtering is a key component in our pipeline. It uses the holistic information derived from the holistic branch to actively filter out noisy pixel predictions in a segmentation map. The idea is to use the holistic information to recommend labels for pixel predictions -- if the holistic information suggests a semantic label is unlikely to be present, then pixels should be unlikely to be predicted as that label as well.

\begin{figure}[tp]
\centering
\includegraphics[width=0.49\textwidth]{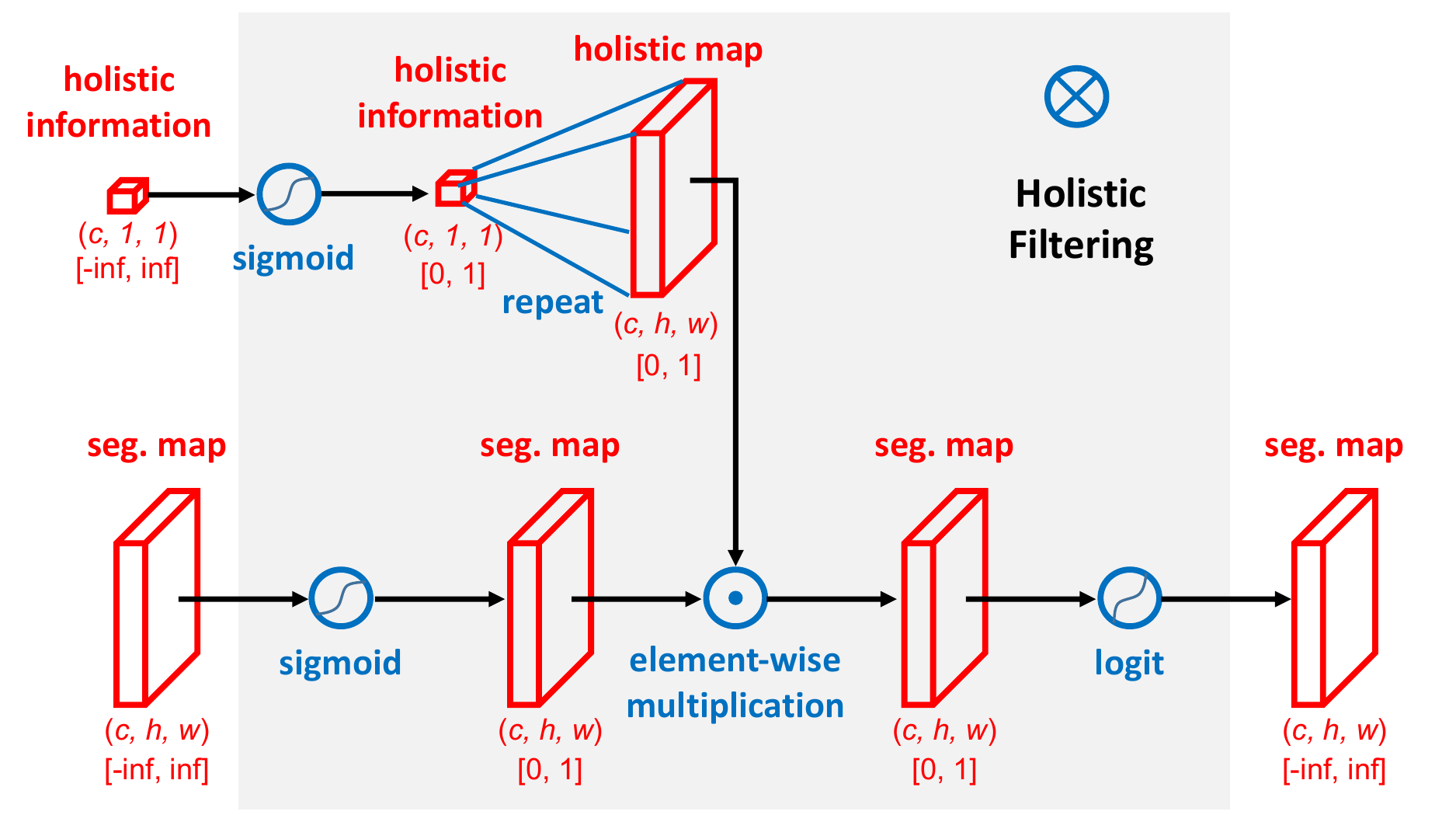}
\caption{Details of the holistic filtering. The notations in round brackets indicate the size of the data, and the numbers in square brackets show that domain of the data values. All operations are differentiable for end-to-end training.}
\label{fig:method:filter}
\vspace{-0.05in}
\end{figure}

Figure~\ref{fig:method:filter} illustrates the detailed implementation of the holistic filtering. Note that the holistic information (\textit{i.e.}, the image label confidences) and the segmentation map are both values in the real domain. Therefore, we first apply a sigmoid function to normalize both to $[0, 1]$ -- a high value indicates a high confidence of observing the corresponding semantic label. Then we multiply the holistic information to each cell of the segmentation map to filter out noisy pixel predictions. In the filtered segmentation map, a cell receives a high value on a label only if both the holistic information and the original pixel prediction are highly confident of predicting that label. Finally, we follow~\cite{hu_cvpr16} to apply a logit function (\textit{i.e.}, the inverse sigmoid function) to map the segmentation map back to the real domain. Note that all operations in the holistic filtering are differentiable so that the gradient can be back-propagated for end-to-end training.

With the filtered segmentation map obtained from holistic filtering, we then apply an up-sampling operation to generate our final output -- a full semantic segmentation map. The up-sampling is simply done by bi-linear interpolation (following~\cite{chen_arxiv16}) to increase the resolution to the original image size. We could also switch the order of up-sampling and holistic filtering in the pipeline to adopt holistic information to filter the full segmentation map. This results in slightly better empirical performance, but increases the computational cost significantly. We keep up-sampling after holistic filtering for the sake of efficiency.

\subsection{Training}
\label{sec:method:train}

We optimize both classification and segmentation losses in our end-to-end trainable network. The classification loss is enforced on the patch network in the holistic branch. Each cell of the classification map corresponds to an image patch, and we impose loss on each patch individually for location-aware recognition. Specifically, we derive a ground-truth classification map from the ground-truth pixel labels. A map cell is labeled as 1 if the corresponding image patch has at least one pixel labeled the same as the map cell's semantic label; and 0 otherwise. We first adopt a sigmoid activation layer on the output classification map, and then evaluate a categorical cross-entropy loss.

The segmentation loss is enforced on the full segmentation map dumped from the segmentation branch. Following the standard setting of semantic segmentation~\cite{shelhamer_pami16,chen_arxiv16}, we first apply a softmax activation layer to the full segmentation map, and then compute a categorical cross-entropy loss for the segmentation task. The balance between the two losses is controlled by a constant multiplier so that both losses have similar order of magnitude.

We follow the practice of FCN~\cite{shelhamer_pami16} to train our network -- optimizing the objective by stochastic gradient descent (SGD) with a small batch size (\textit{e.g.}, 1) and a large momentum (\textit{e.g.}, 0.99). We train our network for approximately 60 epochs and choose the best models through validation. To further clarify our implementation details, we will publicly release our code.

\section{Experiments}
\label{sec:experiment}

We evaluate our network on four benchmark datasets: the PASCAL-Context dataset~\cite{mottaghi_cvpr14}, the ADE20K dataset~\cite{zhou_arxiv16}, the NYUDv2 dataset~\cite{silberman_eccv12}, and the SIFT-Flow dataset~\cite{liu_pami11}. All these datasets contain abundant contextual labels on pixels, and thus are challenging for semantic segmentation tasks. We choose these four datasets as all the pixels are densely labeled with a variety of semantic classes.

\textbf{Baselines}. A direct baseline of our network is to keep the segmentation branch and ignore the holistic branch as well as the holistic filtering operation. As stated in Section~\ref{sec:method:segmentation}, our pixel network in the segmentation branch has two variants modified from FCN~\cite{shelhamer_pami16} and DilatedNet~\cite{yu_iclr16,chen_arxiv16}, so we call the two resultant baselines FCN$^{+}$ and DilatedNet$^{+}$. Following this convention, we name our holistic filtering semantic segmentation networks as Holistic FCN$^{+}$ and Holistic DilatedNet$^{+}$, respectively.

Our feature network adopts the 152-layer ResNet~\cite{he_cvpr16} by default. We have also evaluated VGGNet~\cite{simonyan_iclr15} as an alternative, and the result is in Appendix~\ref{sec:sup:feature}. It shows that our network architecture has flexibility to take various feature networks to improve semantic segmentation.

\textbf{Metrics}. Following~\cite{shelhamer_pami16}, we evaluate four performance metrics that are commonly used for semantic segmentation tasks. The metrics are variations on pixel accuracy and region intersection over union (IU). Specifically, we denote by $n_{ij}$ the number of pixels of class $i$ predicted to belong to class $j$, $t_{i} = \sum_{j} n_{ij}$ the total number of pixels of class $i$, and $c$ the total number of semantic classes. We evaluate:
\vspace{-0.05in}
\begin{itemize}
\setlength{\itemsep}{1pt}
\setlength{\parskip}{1pt}
\setlength{\parsep}{1pt}
\item Pixel Accuracy (pAcc): $\sum_{i} n_{ii} / \sum_{i} t_{i}$
\item Mean Accuracy (mAcc): $(1/c) \sum_{i} n_{ii}/t_{i}$
\item Mean IU (mIU): $(1/c) \sum_{i} n_{ii} / \big(t_{i} + \sum_{j} n_{ji} - n_{ii}\big)$
\item Frequency Weighted IU (fwIU):\\
$(\sum_{k} t_{k})^{-1} \sum_{i} t_{i} n_{ii} / \big(t_{i} + \sum_{j} n_{ji} - n_{ii}\big)$
\end{itemize}
\vspace{-0.05in}
In all the result tables, we highlight the best figures in red and boldfaced, and the second best in blue and underline.

\textbf{Data augmentation}. We describe our data augmentation strategy in Appendix~\ref{sec:sup:aug}.


\textbf{Visualizations}. To smooth paper overflow, we also defer qualitative visualizations to Appendix~\ref{sec:sup:viz}.

\subsection{PASCAL-Context}
\label{sec:experiment:pascal}

\textbf{Dataset}. The PASCAL-Context dataset is derived from the PASCAL VOC 2010 dataset with detailed annotations on every pixel~\cite{mottaghi_cvpr14}. The semantic labels include both objects and stuff present in the image. Following~\cite{mottaghi_cvpr14,shelhamer_pami16}, we evaluate our network on the most frequent 59 classes alongside one background class. The training and testing sets contain 4,998 and 5,105 images, respectively.

\begin{table}[tp]
\small
\tabcolsep 4pt
\centering
\begin{tabular}{ c|cccc }
 & pAcc & mAcc & mIU & fwIU \\
\hlineB{4}
{CFM~\cite{dai_cvpr15}}                         & -           & -           & 34.4            & -           \\ 
{FCN-8s~\cite{shelhamer_pami16}}                & 67.45       & 52.31       & 39.12           & 53.03       \\ 
{CRF-RNN~\cite{zheng_iccv15}}                   & -           & -           & 39.3            & -           \\ 
{DeepLab~\cite{chen_arxiv16}}                   & -           & -           & 39.6            & -           \\ 
{ParseNet~\cite{liu_iclr16}}                    & -           & -           & 40.4            & -           \\ 
{BoxSup~\cite{dai_iccv15}}                      & -           & -           & 40.5            & -           \\ 
{HO-CRF~\cite{arnab_eccv16}}                    & -           & -           & 41.3            & -           \\ 
{Context~\cite{lin_cvpr16}}                     & 71.5        & 53.9        & 43.3            & -           \\ 
{DeepLab + COCO~\cite{chen_arxiv16}}            & -           & -           & 44.7            & -           \\ 
{DeepLab + COCO + CRF~\cite{chen_arxiv16}}      & -           & -           & 45.7            & -           \\ 
\hline
{FCN$^{+}$}                                  & 71.25       & 53.82       & 43.19           & 57.63          \\ 
{Holistic FCN$^{+}$}                         & \bst{73.52} & \bst{56.72} & \second{45.77}  & \second{60.05} \\
{DilatedNet$^{+}$}                           & 70.26       & 53.10       & 41.72           & 56.57          \\ 
{Holistic DilatedNet$^{+}$}                  & \second{73.47} & \second{56.59} & \bst{45.80} & \bst{60.23}  \\
\hline
\end{tabular}
\caption{Semantic segmentation results on PASCAL-Context.}
\label{tab:experiment:pascal}
\end{table}

\textbf{Evaluation}. We have compared our holistic filtering networks with our baselines and existing methods in the literature. The evaluation results are reported in Table~\ref{tab:experiment:pascal}. It clearly shows that the two holistic filtering networks outperform all the other methods, in terms of all four performance metrics. This verifies the effectiveness of the proposed method.

It is worth noting that both Holistic FCN$^{+}$ and Holistic DilatedNet$^{+}$ outperform the current state-of-the-art method, DeepLab + COCO + CRF~\cite{chen_arxiv16}. However, we did not use extra training data (\textit{e.g.} the COCO dataset) and domain adaptation to obtain a better model. We applied neither CRFs to smooth the segmentation results, nor multi-scale test to refine segmentation at various image granularities. Therefore, we suspect that our network performance can be further boosted once we apply these techniques.

Also note that holistic filtering assists in semantic segmentation. For example, the mIU values of FCN$^{+}$ and DilatedNet$^{+}$ is improved by 2.85\% and 3.51\% respectively with holistic filtering. It shows the effectiveness of our two-stream network architecture that recalls holistic information for semantic segmentation.

\subsection{ADE20K}
\label{sec:experiment:ade}

\textbf{Dataset}. ADE20K is a recently released dataset with densely annotated objects and stuff. We learn our model on the training set of 20,210 images, and report performance on the 2,000 validation images. We did not evaluate on the 5,000 test images as the ground-truth annotations have not been made public yet. Following~\cite{zhou_arxiv16}, we select the top 150 semantic classes ranked by their total pixel ratios, including 35 object classes and 115 stuff classes. The pixels from the 150 classes occupy 92.75\% of all pixels in the dataset.

\begin{table}[tp]
\small
\tabcolsep 4pt
\centering
\begin{tabular}{ c|cccc }
 & pAcc & mAcc & mIU & fwIU \\
\hlineB{4}
{FCN-8s~\cite{zhou_arxiv16}}                & 71.32       & 40.32       & 29.39       & 57.33           \\ 
{DilatedNet~\cite{zhou_arxiv16}}            & 73.55       & 44.59       & 32.31       & 60.14           \\ 
{DilatedNet Cascade~\cite{zhou_arxiv16}}    & 74.52       & 45.38       & 34.90       & 61.08           \\ 
\hline
{FCN$^{+}$}                              & 77.30         & 46.94        & 36.52       & 64.49             \\ 
{Holistic FCN$^{+}$}                     & \second{78.03} & \bst{48.23}  & \bst{37.93} & \second{65.34}   \\
{DilatedNet$^{+}$}                       & 76.63         & 44.83        & 34.15       & 63.68             \\
{Holistic DilatedNet$^{+}$}              & \bst{78.20}   & \second{47.89}  & \second{37.66} & \bst{65.62} \\
\hline
\end{tabular}
\caption{Semantic segmentation results on the ADE20K validation set, using 384$\times$384 training and testing images.}
\label{tab:experiment:ade-384}
\end{table}

\begin{table}[tp]
\small
\tabcolsep 2.5pt
\centering
\begin{tabular}{ c|cccc }
 & pAcc & mAcc & mIU & fwIU \\
\hlineB{4}

{ILSVRC 2016 Best w. ResNet-269}    & {80.90}     & -     & {43.39}     & -              \\
{ILSVRC 2016 Best w. ResNet-152}    & {80.46}     & -     & {42.23}     & -              \\
\hline
{FCN$^{+}$}                          & {77.64}     & {47.11}    & {37.20}   & {64.85}       \\ 
{Holistic FCN$^{+}$}                 & {79.15}     & {51.61}    & {40.50}   & {66.82}       \\
\hline
\end{tabular}
\caption{Semantic segmentation results on the ADE20K validation set, using full-resolution images for training and testing. We compare with the ILSVRC 2016 best entry~\cite{zhao_slides16} using single model and single-scale test.}
\label{tab:experiment:ade-ilsvrc}
\end{table}

\textbf{Evaluation}. We perform two sets of experiments. The first set follows~\cite{zhou_arxiv16}. We train and test our networks on re-sized images of 384 by 384 pixels. To evaluate the performance against the ground-truth annotations, we re-size the resultant segmentation on each test image from 384$\times$384 back to the original image size. This experiment enables us to directly compare with those reported in~\cite{zhou_arxiv16}, and we apply this setting by default for ADE20K experiments if not further specified. Note that this setting may lead to the loss of image details and object aspect ratios, thus yielding sub-optimal performance.

We summarize the results in Table~\ref{tab:experiment:ade-384}. Again it shows the utility of holistic filtering networks, which achieve the best performance in all four metrics. The results also verify the usage of holistic filtering -- it boosts FCN$^{+}$ and DilatedNet$^{+}$ remarkably.

The second set of experiments follows the ILSVRC 2016 Scene Parsing Challenge, and uses full-resolution images for both training and testing. We compare our Holistic FCN$^{+}$ with the ILSVRC 2016 best entry~\cite{zhao_slides16} (under single model and single-scale test) in Table~\ref{tab:experiment:ade-ilsvrc}. Holistic DilatedNet$^{+}$ is not included due to the intensive dilated convolution on full-resolution images.


Table~\ref{tab:experiment:ade-ilsvrc} show that that our method is competitive. Specifically, the mIU of Holistic FCN$^{+}$ is 1.73\% behind the best result with the same feature network (the 152-layer ResNet), and 2.89\% behind the best entry (built on the 269-layer ResNet). Note that we did not investigate all practical techniques to further improve our network performance, such as leveraging extra scene labels, applying deeply supervised training, pre-training with PlaceNet, using a deeper CNN architecture as the feature network, \textit{etc}.





\subsection{NYUDv2}
\label{sec:experiment:nyud}

\textbf{Dataset}. NYUDv2 is an RGB-D dataset on indoor scenes collected using Microsoft Kinect. It has 1,449 RGB-D images, with pixel-wise labels that have been coalesced into 40 semantic classes by Gupta \textit{et al.}~\cite{gupta_cvpr13}. We experiment with the standard split of 795 training images and 654 testing images.

\textbf{Evaluation}. We compare our methods with state-of-the-art approaches, and report the evaluation results in Table~\ref{tab:experiment:nyud}. Our holistic filtering networks achieve the best performance, especially with Holistic FCN$^{+}$.

\begin{table}[tp]
\small
\tabcolsep 4pt
\centering
\begin{tabular}{ c|cccc }
 & pAcc & mAcc & mIU & fwIU \\
\hlineB{4}
{Gupta~\textit{et al.}~\cite{gupta_eccv14}}         & 60.3          & -         & 28.6      & 47.0          \\
{FCN-32s + RGB~\cite{shelhamer_pami16}}             & 61.8          & 44.7      & 31.6      & 46.0          \\ 
{FCN-32s + RGB + D~\cite{shelhamer_pami16}}         & 62.1          & 44.8      & 31.7      & 46.3          \\ 
{FCN-32s + RGB + HHA~\cite{shelhamer_pami16}}       & \bst{65.3}    & 44.0      & 33.3      & 48.6          \\ 
\hline
{FCN$^{+}$}                           & {64.60}     & {48.40}           & {37.19}         & \second{49.41}  \\ 
{Holistic FCN$^{+}$}                  & \bst{65.32} & \bst{50.82}       & \bst{38.76}     & \bst{50.25}     \\ 
{DilatedNet$^{+}$}                    & 62.88       & 46.13             & 34.74           & 47.64           \\
{Holistic DilatedNet$^{+}$}           & {64.14}     & \second{49.67}    & \second{37.23}  & {49.08}         \\ 
\hline
\end{tabular}
\caption{Semantic segmentation results on NYUDv2.}
\label{tab:experiment:nyud}
\end{table}

Our networks only use the color images for training, ignoring the depth information. We still improve 5.46\% mIU over the state-of-the-art method, FCN-32s + RGB + HHA~\cite{shelhamer_pami16}, which utilizes both the color and depth information. It again verifies the effectiveness of holistic filtering and our two-stream network architecture.

\subsection{SIFT-Flow}
\label{sec:experiment:siftflow}

\textbf{Dataset}. The SIFT-Flow dataset contains 2,688 images thoroughly annotated by LabelMe users~\cite{liu_pami11,liu_pami11a} with 33 semantic pixel labels (\textit{e.g.}, mountain, sun, bridge, \textit{etc}). We use the same split as~\cite{liu_pami11,liu_pami11a} -- 2,488 images for training and 200 images for testing.

\textbf{Evaluation}. We compare our networks with state-of-the-art approaches in Table~\ref{tab:experiment:siftflow}. The results show that our holistic filtering networks perform the best over all the compared methods. Note that current state-of-the-art FCN-8s~\cite{shelhamer_pami16} leverages the available pixel-wise geometric labels (\textit{i.e.}, horizontal, vertical and sky) as extra supervision, whereas our networks do not use them. These observations show the utility of our networks in semantic segmentation.

\begin{table}[tp]
\small
\tabcolsep 3pt
\centering
\begin{tabular}{ c|cccc }
 & pAcc & mAcc & mIU & fwIU \\
\hlineB{4}
{Liu~\textit{et al.}~\cite{liu_pami11a}}    & 76.7        & -           & -                 & -      \\
{Exemplar SVM~\cite{tighe_cvpr13}}          & 75.6        & 41.1        & -                 & -      \\
{SVM + MRF~\cite{tighe_cvpr13}}             & 78.6        & 39.2        & -                 & -      \\
{Multiscale CNN + Natural~\cite{farabet_pami13}} & 72.3   & 50.8        & -                 & -      \\
{Multiscale CNN + Balanced~\cite{farabet_pami13}} & 78.5  & 29.6        & -                 & -      \\
{Recurrent CNN~\cite{pinheiro_icml14}}      & 77.7        & 29.8        & -                 & -      \\
{ParseNet~\cite{liu_iclr16}}                & 86.8        & 52.0        & 40.4              & 78.1   \\
{FCN-8s~\cite{shelhamer_pami16}}            & 85.9        & 53.9        & 41.2              & 77.2   \\
\hline
{FCN$^{+}$}                                 & {87.90}        & {53.04}          & {42.19}        & {79.80}         \\
{Holistic FCN$^{+}$}                        & \bst{88.16}    & \second{55.92}   & {42.11}        & \bst{80.51}     \\
{DilatedNet$^{+}$}                          & {87.23}        & {55.32}          & \second{42.24} & \second{80.27}     \\
{Holistic DilatedNet$^{+}$}                 & \second{88.01} & \bst{58.62}      & \bst{44.14}    & {80.21}     \\
\hline
\end{tabular}
\caption{Semantic segmentation results on SIFT-Flow.}
\label{tab:experiment:siftflow}
\end{table}

It is worth mentioning that a competitive method on the SIFT-Flow dataset is proposed by Lin~\textit{et al.}~\cite{lin_cvpr16}, achieving 88.1\% pAcc, 53.4\% mAcc and 44.9\% mIU. However, this method up-samples training and testing images by a factor of two, and benefits from the high-resolution images to obtain superior performance. On the other hand, all other methods including our networks do not up-sample images, and thus are not directly comparable with~\cite{lin_cvpr16}.

\subsection{Analysis}
\label{sec:experiment:analysis}

All the above experiments show that semantic segmentation benefits from holistic information and holistic filtering. However, by no means would we claim that our holistic branch is perfect. For example, when applying a threshold of 0 on the image label confidences from the holistic information derived by Holistic FCN$^{+}$, we observe 90.93\% recall and 46.75\% precision on PASCAL-Context, and 69.07\% recall and 47.59\% precision on ADE20K. Thus, an interesting question to raise is: \textit{How well our networks could perform if we have a perfect holistic branch with 100.00\% recall and 100.00\% precision?}

We answer this question by leveraging ground-truth image labels in Holistic FCN$^{+}$. We implement it by replacing our holistic branch with a static process that returns ground-truth image label confidences, taking the value of infinity if a label is present in the image; and negative infinity otherwise. We keep the rest of the network the same as Holistic FCN$^{+}$, and train it end-to-end towards optimizing the segmentation loss (see Section~\ref{sec:method:train} for details of the loss).

\begin{figure}[tp]
\begin{center}
\includegraphics[width=0.495\textwidth]{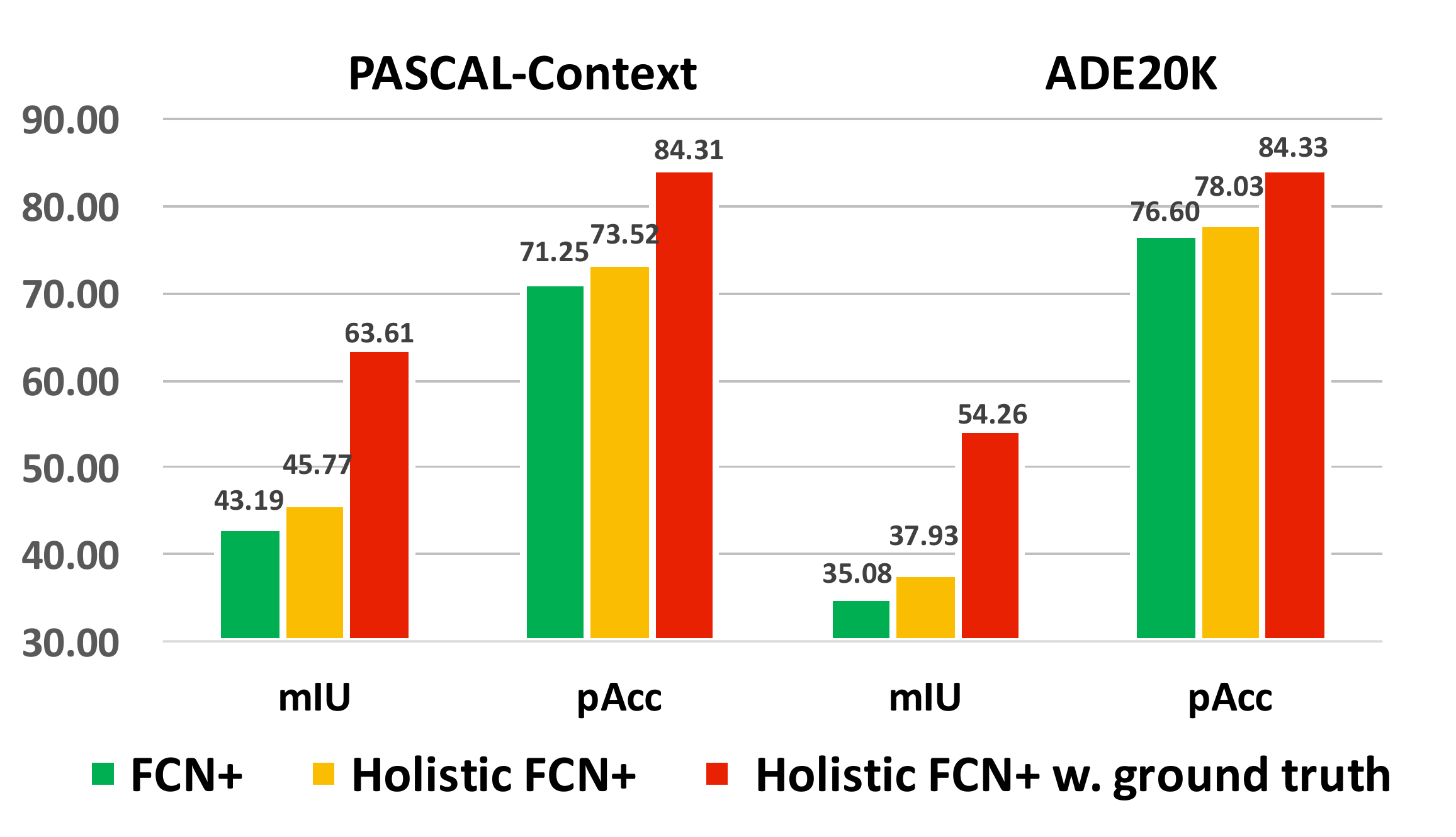}
\end{center}
\caption{The pAcc and mIU performance of FCN$^{+}$, Holistic FCN$^{+}$ and Holisitic FCN$^{+}$ with ground truth on PASCAL-Context and ADE20K.}
\label{fig:experiment:analysis}
\end{figure}

We compare the pACC and mIU performance of FCN$^{+}$, Holistic FCN$^{+}$, and Holistic FCN$^{+}$ with ground truth in Figure~\ref{fig:experiment:analysis}. It clearly shows that Holistic FCN$^{+}$ with ground truth significantly boosts semantic segmentation -- for example, the mIU values are improved to 63.61\% on PASCAL-Context and 54.26\% on ADE20K. We believe Holistic FCN$^{+}$ with ground truth provides the upper-bound on performance we could have achieved with holistic filtering, and we would gradually approach this upper-bound as we obtain a better and better holistic analysis models in our two-stream network architecture. This opens up a promising direction for further research.

\begin{figure*}[tp]
\centering
\includegraphics[width=1\textwidth]{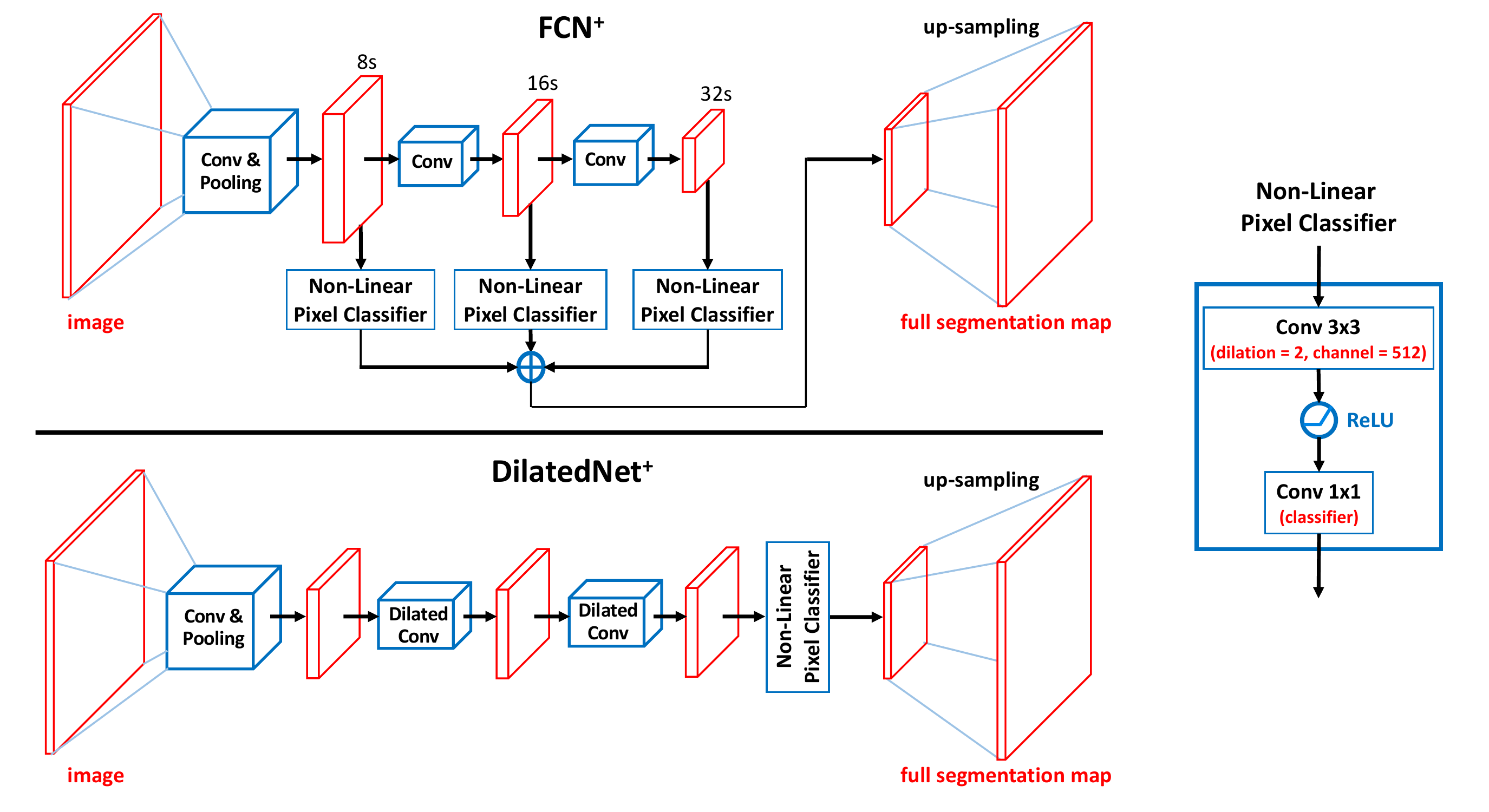}
\caption{The network architectures of FCN$^{+}$ and DilatedNet$^{+}$. Note that the only modification from FCN~\cite{shelhamer_pami16} and DilatedNet~\cite{chen_arxiv16} is to use our non-linear pixel classifier to generate segmentation maps. More details are described in the text.}
\label{fig:sup:network:plus}
\end{figure*}

\section{Conclusion}
\label{sec:conclusion}

This paper is motivated by the benefit of holistic information in semantic segmentation. We have presented empirical analysis to support our intuition, and proposed a two-stream neural network architecture as an implementation. Our network includes a holistic branch to recall holistic information by inspecting location-aware image patches. We also have a segmentation branch that leverages holistic filtering to filter out noisy pixel predictions. Experiments on four benchmark datasets show the effectiveness of the proposed method, which achieves state-of-art performance on PASCAL-Context and NYUDv2, as well as competitive results on ADE20K and SIFT-Flow. Our method that leverages holistic information from images is general and could be applied to many other applications, for example, improving object detection with holistic image labels, locating actions with holistic video activity annotations, and so on.

\begin{appendices}

\section{FCN$^+$ and DilatedNet$^+$}
\label{sec:sup:plus}

As mentioned in Section~\ref{sec:method:segmentation}, the pixel network in our segmentation branch is responsible for generating a segmentation map based on the feature map obtained from the feature network. We have implemented two variants as our pixel network in our segmentation branch. Specifically, we have FCN$^{+}$ derived from FCN~\cite{shelhamer_pami16} and DilatedNet$^{+}$ derived from DilatedNet~\cite{chen_arxiv16}. We show the network architectures in Figure~\ref{fig:sup:network:plus}, and describe the details in the following.

FCN$^{+}$ has a similar structure as FCN~\cite{shelhamer_pami16}: it first uses skip features to generate three segmentation maps of various down-sampling rates (8s, 16s and 32s), then fuses them together, and finally up-samples to the original image size for semantic segmentation. The only modification is that FCN$^{+}$ applies our designed \textit{non-linear pixel classifier} to generate a segmentation map from the input feature map.

The non-linear pixel classifier leverages non-linearity and dilated convolution to model pixel labeling. It consists of a dilated convolutional layer~\cite{yu_iclr16,chen_arxiv16} to summarize image information, a ReLU layer for non-linearity, as well as a one-by-one convolutional layer to predict the segmentation map. In our experiments, we apply 3 by 3 convolutional kernels in the dilated convolutional layer, using a dilation rate of 2 under 512 channels.

Our DilatedNet$^{+}$ is derived from DilatedNet~\cite{chen_arxiv16} with the same modification -- we apply our non-linear pixel classifier on the feature map (obtained after dilated convolutions) to generate the segmentation map. The rest of the structure is kept the same as~\cite{chen_arxiv16}.

Note that our experimental results verify the utility of our non-linear pixel classifier. See Tables~\ref{tab:experiment:pascal}, \ref{tab:experiment:ade-384}, \ref{tab:experiment:nyud}, and \ref{tab:experiment:siftflow} for detailed comparison between FCN$^{+}$ and FCN, and between DilatedNet$^{+}$ and DilatedNet.

\begin{table*}[tp]
\small
\tabcolsep 15pt
\centering
\begin{tabular}{ c|c|cccc }
 & Feature Network & pAcc & mAcc & mIU & fwIU \\ \hlineB{4} 
{FCN$^{+}$}                 & VGGNet-16     & 68.64 & 52.36 & 39.45 & 54.50 \\ 
{Holistic FCN$^{+}$}        & VGGNet-16     & 69.24 & 51.81 & 40.11 & 55.14 \\
{DilatedNet$^{+}$}          & VGGNet-16     & 68.28 & 52.30 & 39.21 & 53.75 \\ 
{Holistic DilatedNet$^{+}$} & VGGNet-16     & 70.50 & 54.24 & 41.83 & 56.37 \\ \hlineB{4}
{FCN$^{+}$}                 & ResNet-152    & 71.25 & 53.82 & 43.19 & 57.63 \\
{Holistic FCN$^{+}$}        & ResNet-152    & \bst{73.52} & \bst{56.72} & \second{45.77} & \second{60.05} \\
{DilatedNet$^{+}$}          & ResNet-152    & 70.26 & 53.10 & 41.72 & 56.57 \\ 
{Holistic DilatedNet$^{+}$} & ResNet-152    & \second{73.47} & \second{56.59} & \bst{45.80} & \bst{60.23} \\
\hline
\end{tabular}
\caption{Ablation study of feature networks on the PASCAL-Context dataset~\cite{mottaghi_cvpr14}. We highlight the best performance in red and boldfaced, and the second best in blue and underline.}
\label{tab:sup:feature:pascal}
\end{table*}

\section{Precision and Recall Setup}
\label{sec:sup:pr}

In Section~\ref{sec:analysis:recall}, we experimented with the holistic filtering methods with various precision and recall settings of ground-truth labels in holistic information. The detailed experimental setup is as follows.

To degrade the precision, we have added in $n_{p} \in \{1, 2, \cdots, 10\}$ noisy labels per image. Similarly, to degrade the recall, we have removed $n_{r} \in \{1, 2, \cdots, 10\}$ ground-truth labels per image (if there are fewer than $n_{r}$ ground-truth labels in an image, we just remove them all). Note that $n_{p}$ and $n_{r}$ can be set as fractional values to further refine the numerical accuracy of precision and recall. For example, setting $n_{r}=2.3$ means that we will first remove 2 (the integer part of $n_{r}$) ground-truth labels on each image, and then randomly pick up 30\% (the decimal part of $n_{r}$) of the images to remove one more ground-truth label each. With this trick, we also tried $n_{r} \in \{0.2, 0.4, 0.6, 0.8\}$ in our experiment. We conduct holistic filtering for semantic segmentation with an exhaustive grid search of $n_{p}$ and $n_{r}$ values. Note that each combination of $n_{p}$ and $n_{r}$ produces holistic information with certain precision and recall of ground-truth labels (by averaging over all images). We plot the mean IU performance in Figure~\ref{fig:analysis:grid}.

\section{Data Augmentation Strategy}
\label{sec:sup:aug}

It has been shown that data augmentation is a practical technique to boost semantic segmentation performance. In our experiments, we have applied horizontal flipping as well as scale augmentation when training our networks. For scale augmentation, we randomly pick a scale factor in the set $\{0.7, 0.8, 0.9, 1.0, 1.1, 1.2, 1.3\}$ to scale each image.

\section{Feature Network Ablation Study}
\label{sec:sup:feature}


As promised in Section~\ref{sec:experiment}, we have conducted experiments to study the flexibility of our network architecture in taking various feature networks. In detail, we have tried the 16-layer VGGNet~\cite{simonyan_iclr15} and the 152-layer ResNet~\cite{he_cvpr16} respectively as the feature network, by removing the top fully-connected layers and keeping the convolutional and pooling layers. We evaluate our networks (\textit{i.e.}, Holistic FCN$^{+}$ and Holistic DilatedNet$^{+}$) and our baselines (\textit{i.e.}, FCN$^{+}$ and DilatedNet$^{+}$) using the two feature network variants. The comparison results on the PASCAL-Context~\cite{mottaghi_cvpr14} dataset is reported in Table~\ref{tab:sup:feature:pascal}.

The table clearly shows that our holistic filtering networks improve over the baselines, using either VGGNet or ResNet. It verifies that our network architecture can flexibly adopt various feature networks to improve semantic segmentation. This flexibility enables our network to take advantage of the continual improvement in CNN architectures. Furthermore, it is beneficial to use ResNet as our feature network as it consistently outperforms the VGGNet counterpart. It is reasonable since ResNet employs a deeper structure than VGGNet to capture rich image features.

\section{Visualizations}
\label{sec:sup:viz}

We select sample images from PASCAL-Context and ADE20K, and visualize the semantic segmentation results in Figures~\ref{fig:sup:viz:pascal} and \ref{fig:sup:viz:ade}. As a comparison with existing methods, we have also provided the FCN~\cite{shelhamer_pami16} results on PASCAL-Context and the DilatedNet~\cite{zhou_arxiv16} results on ADE20K. The qualitative results verify the usage of holistic filtering in semantic segmentation -- recalling holistic information to filter out noisy pixel predictions is a practical strategy for semantic segmentation.


\def\VizImageWidth{0.835in}
\def\VizImageHeightLong{1.2in}
\def\VizImageHeightMiddle{0.835in}
\def\VizImageHeightShort{0.7in}

\begin{figure*}[tp]
\small
\centering
\tabcolsep 1pt
\begin{tabular}{cccc|cccc}
Image & FCN~\cite{shelhamer_pami16} & Holistic FCN$^{+}$ & Ground Truth & Image & FCN~\cite{shelhamer_pami16} & Holistic FCN$^{+}$ & Ground Truth \\
\includegraphics[width=\VizImageWidth,height=\VizImageHeightLong]{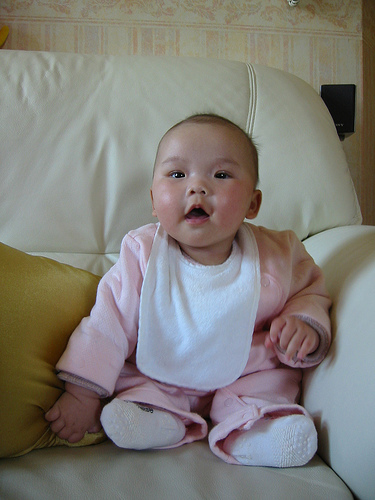} &
\includegraphics[width=\VizImageWidth,height=\VizImageHeightLong]{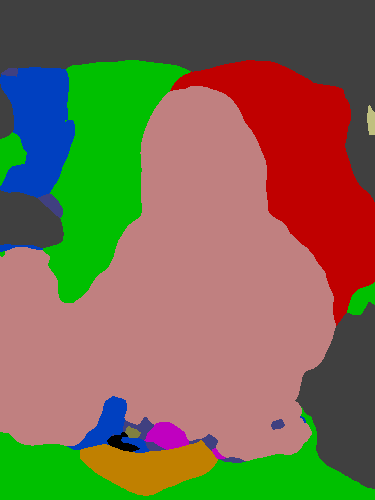} &
\includegraphics[width=\VizImageWidth,height=\VizImageHeightLong]{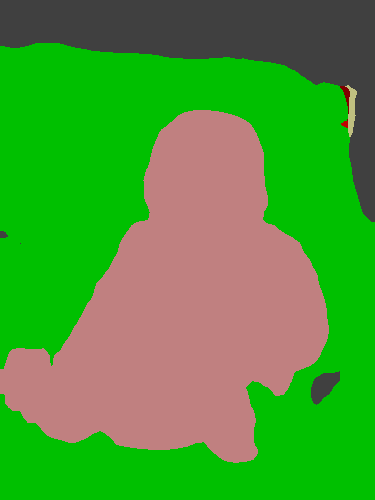} &
\includegraphics[width=\VizImageWidth,height=\VizImageHeightLong]{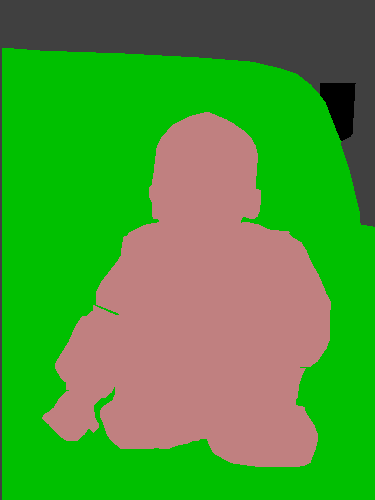} &
\includegraphics[width=\VizImageWidth,height=\VizImageHeightLong]{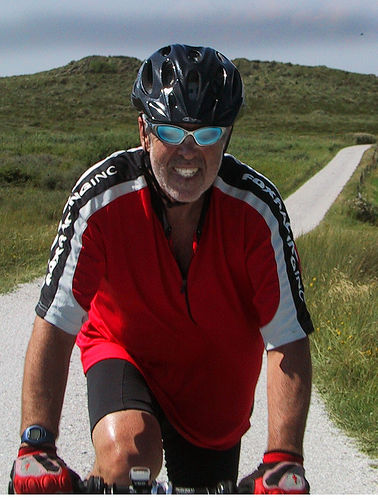} &
\includegraphics[width=\VizImageWidth,height=\VizImageHeightLong]{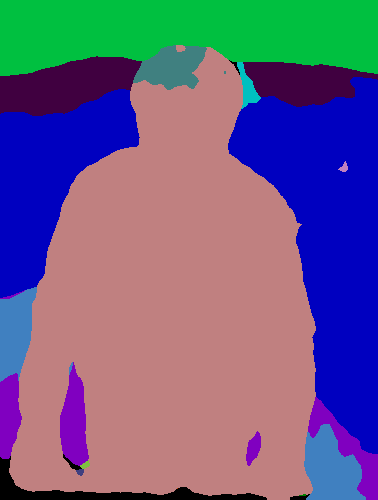} &
\includegraphics[width=\VizImageWidth,height=\VizImageHeightLong]{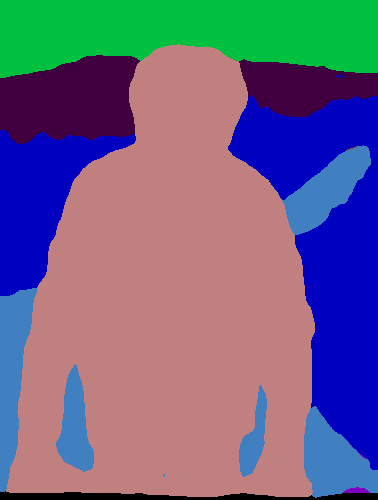} &
\includegraphics[width=\VizImageWidth,height=\VizImageHeightLong]{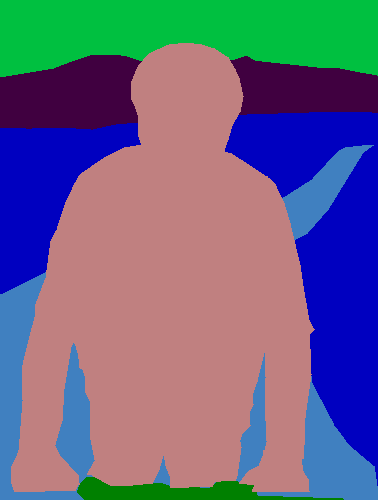} \\
\includegraphics[width=\VizImageWidth,height=\VizImageHeightLong]{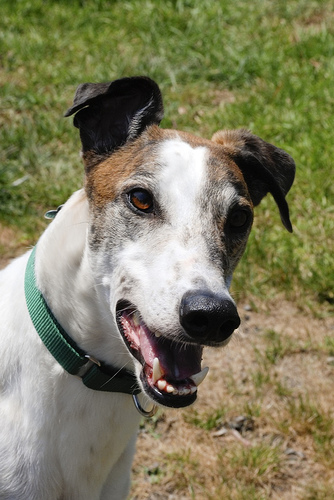} &
\includegraphics[width=\VizImageWidth,height=\VizImageHeightLong]{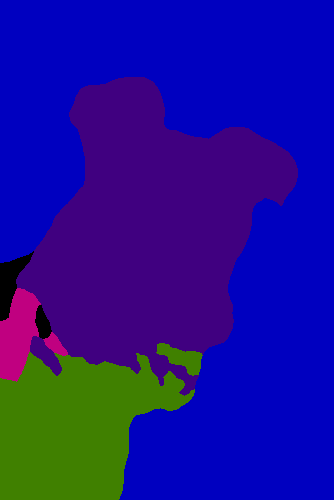} &
\includegraphics[width=\VizImageWidth,height=\VizImageHeightLong]{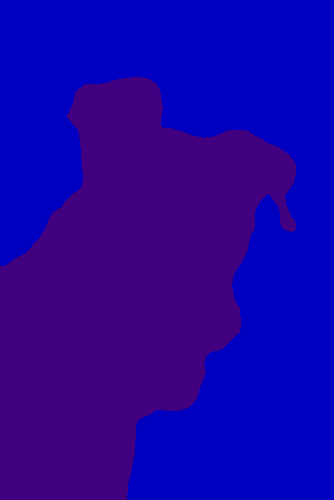} &
\includegraphics[width=\VizImageWidth,height=\VizImageHeightLong]{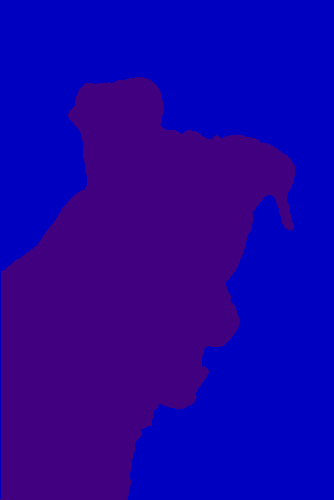} &
\includegraphics[width=\VizImageWidth,height=\VizImageHeightLong]{file/pascal/2008_002383.jpg} &
\includegraphics[width=\VizImageWidth,height=\VizImageHeightLong]{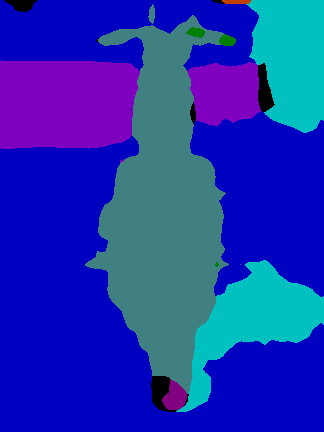} &
\includegraphics[width=\VizImageWidth,height=\VizImageHeightLong]{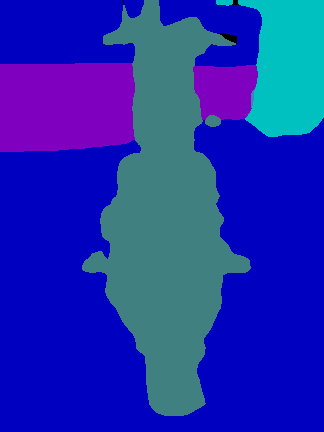} &
\includegraphics[width=\VizImageWidth,height=\VizImageHeightLong]{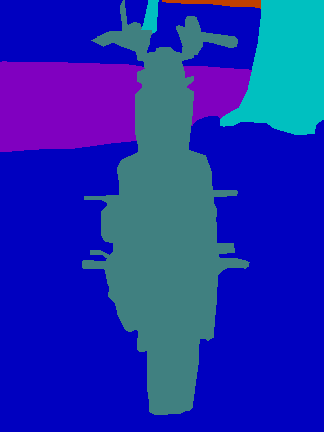} \\
\includegraphics[width=\VizImageWidth,height=\VizImageHeightShort]{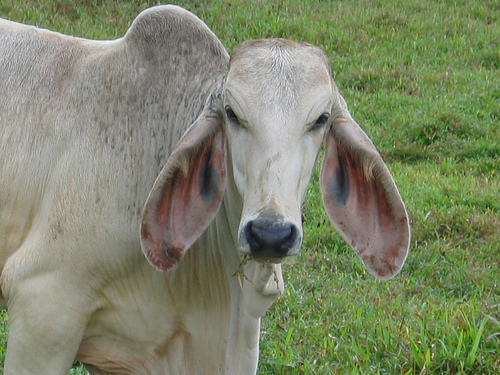} &
\includegraphics[width=\VizImageWidth,height=\VizImageHeightShort]{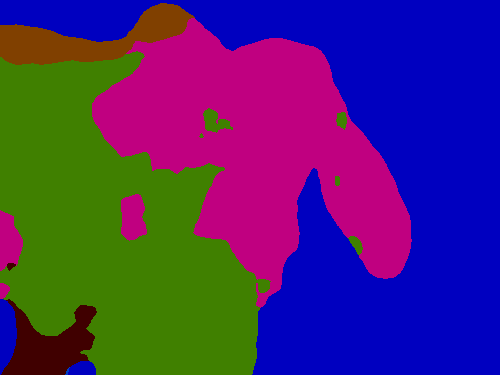} &
\includegraphics[width=\VizImageWidth,height=\VizImageHeightShort]{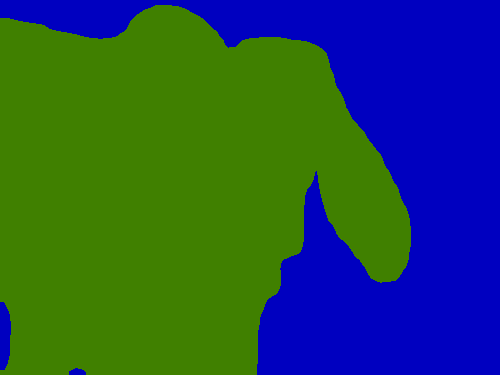} &
\includegraphics[width=\VizImageWidth,height=\VizImageHeightShort]{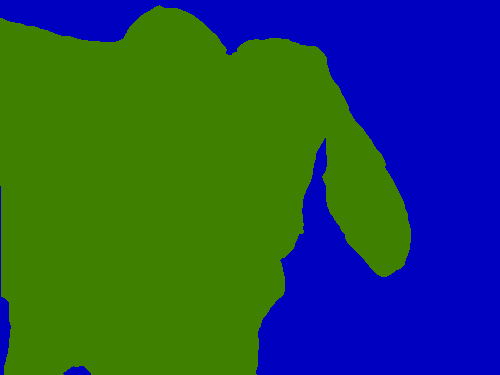} &
\includegraphics[width=\VizImageWidth,height=\VizImageHeightShort]{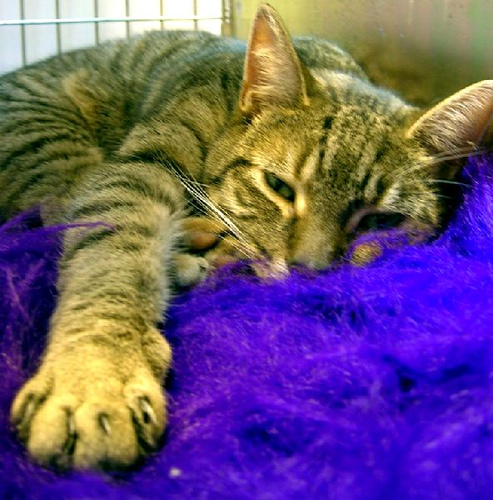} &
\includegraphics[width=\VizImageWidth,height=\VizImageHeightShort]{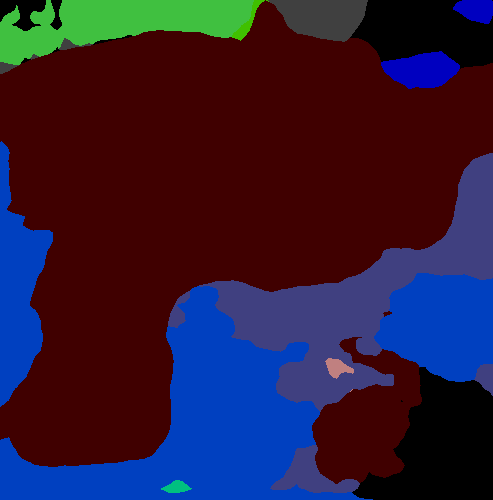} &
\includegraphics[width=\VizImageWidth,height=\VizImageHeightShort]{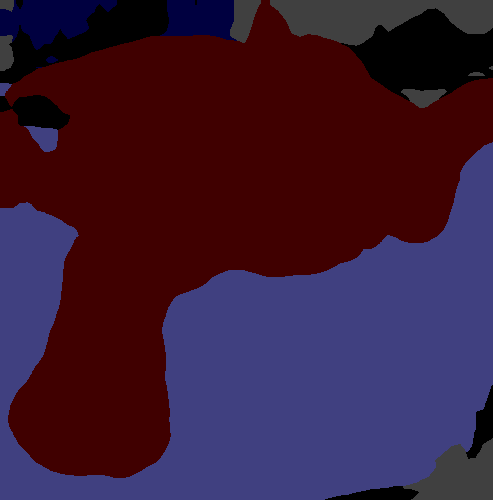} &
\includegraphics[width=\VizImageWidth,height=\VizImageHeightShort]{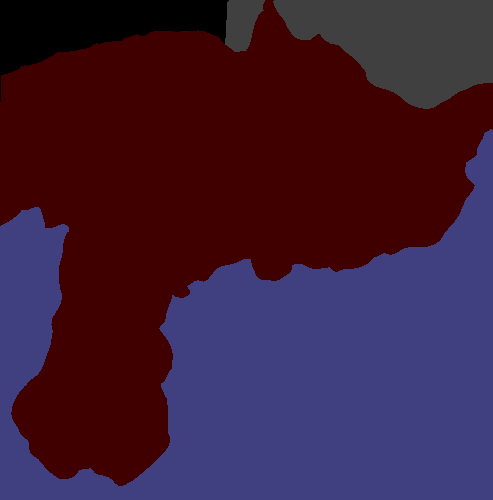} \\
\includegraphics[width=\VizImageWidth,height=\VizImageHeightShort]{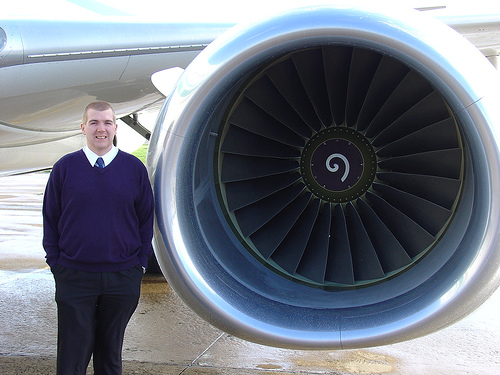} &
\includegraphics[width=\VizImageWidth,height=\VizImageHeightShort]{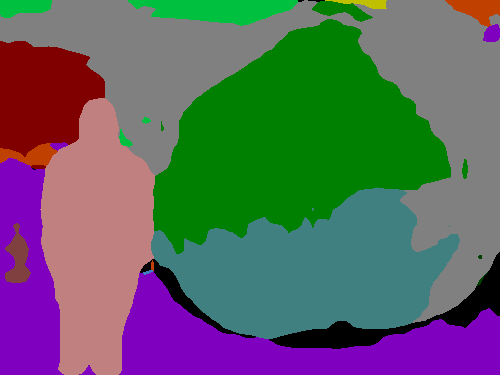} &
\includegraphics[width=\VizImageWidth,height=\VizImageHeightShort]{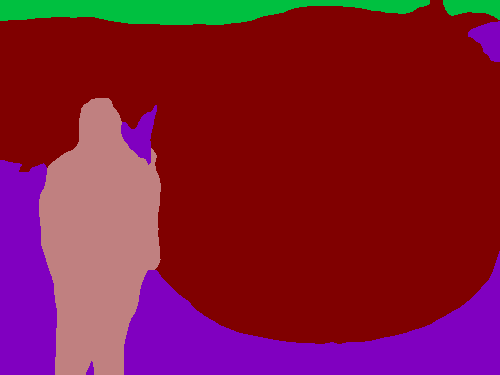} &
\includegraphics[width=\VizImageWidth,height=\VizImageHeightShort]{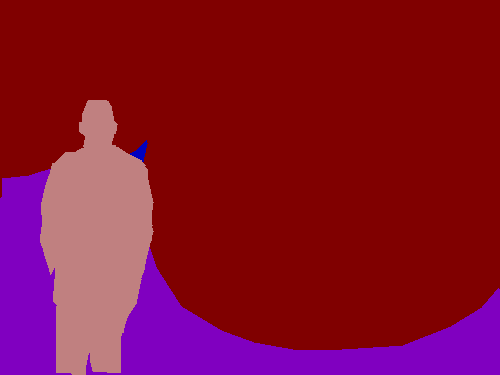} &
\includegraphics[width=\VizImageWidth,height=\VizImageHeightShort]{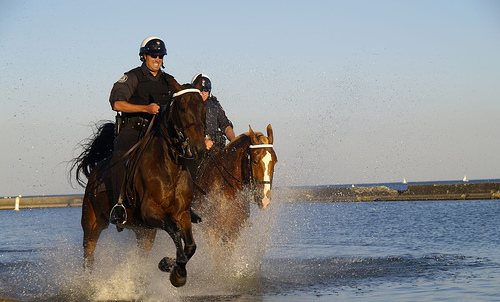} &
\includegraphics[width=\VizImageWidth,height=\VizImageHeightShort]{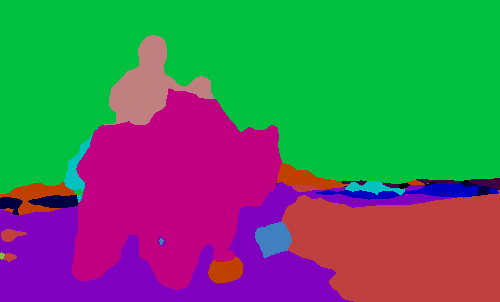} &
\includegraphics[width=\VizImageWidth,height=\VizImageHeightShort]{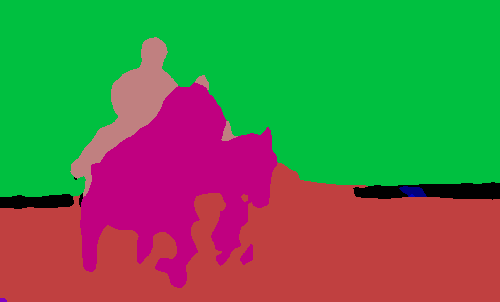} &
\includegraphics[width=\VizImageWidth,height=\VizImageHeightShort]{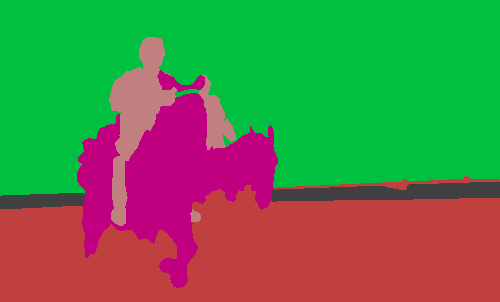} \\
\includegraphics[width=\VizImageWidth,height=\VizImageHeightShort]{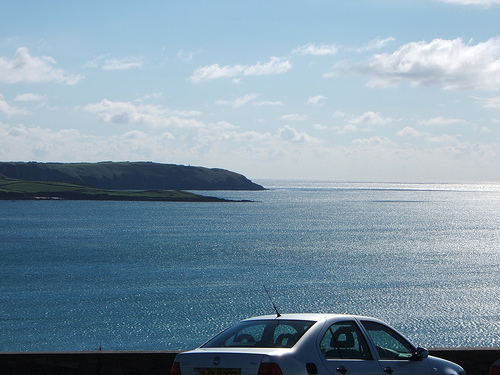} &
\includegraphics[width=\VizImageWidth,height=\VizImageHeightShort]{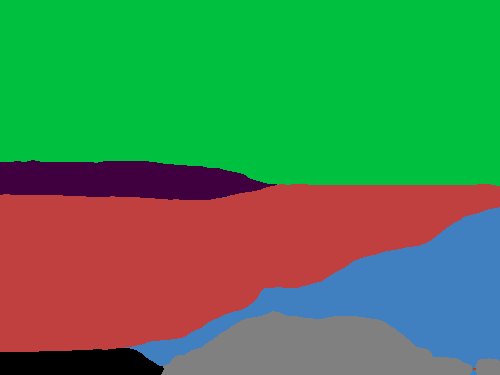} &
\includegraphics[width=\VizImageWidth,height=\VizImageHeightShort]{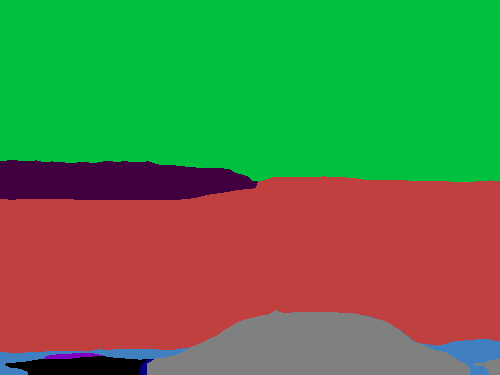} &
\includegraphics[width=\VizImageWidth,height=\VizImageHeightShort]{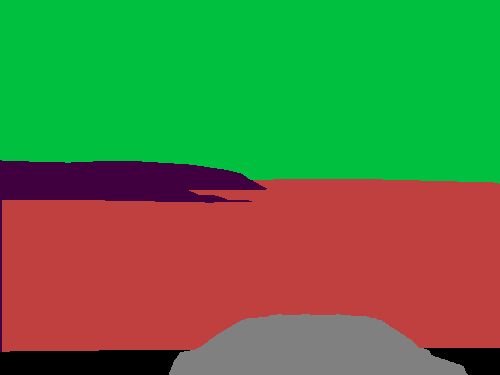} &
\includegraphics[width=\VizImageWidth,height=\VizImageHeightShort]{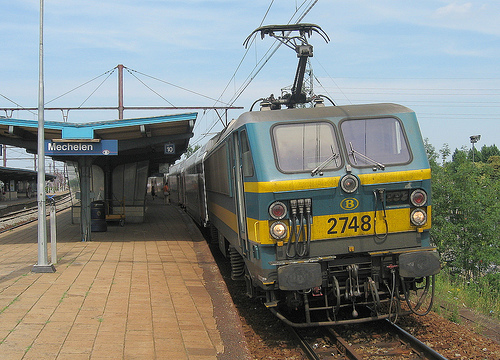} &
\includegraphics[width=\VizImageWidth,height=\VizImageHeightShort]{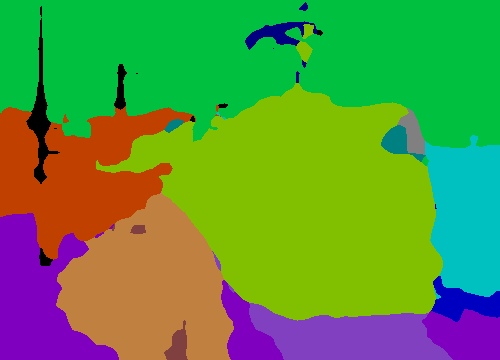} &
\includegraphics[width=\VizImageWidth,height=\VizImageHeightShort]{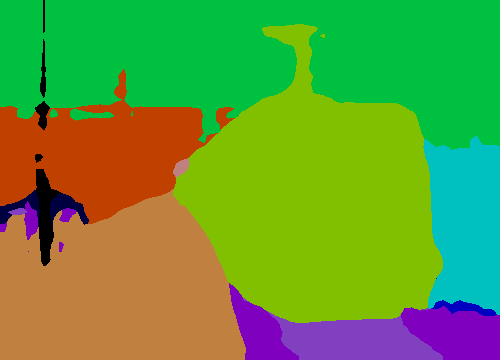} &
\includegraphics[width=\VizImageWidth,height=\VizImageHeightShort]{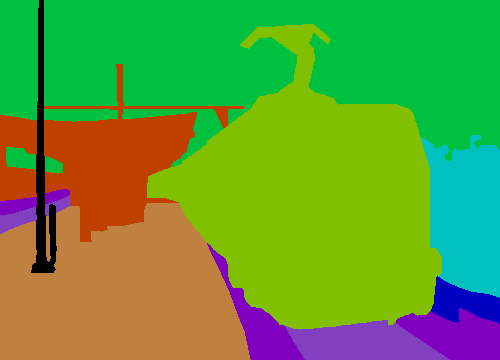} \\
\includegraphics[width=\VizImageWidth,height=\VizImageHeightShort]{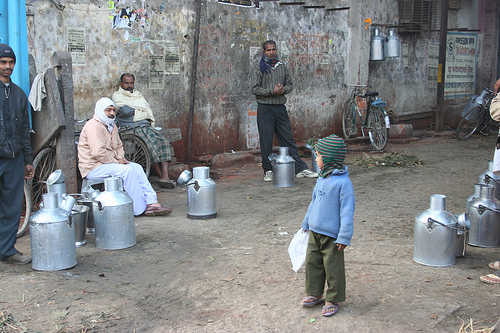} &
\includegraphics[width=\VizImageWidth,height=\VizImageHeightShort]{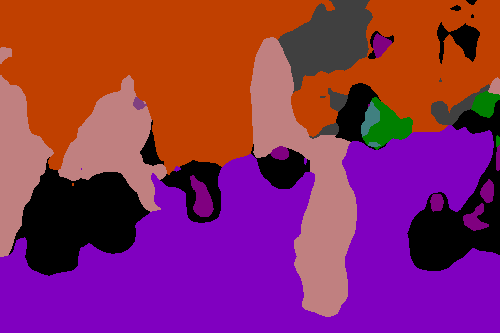} &
\includegraphics[width=\VizImageWidth,height=\VizImageHeightShort]{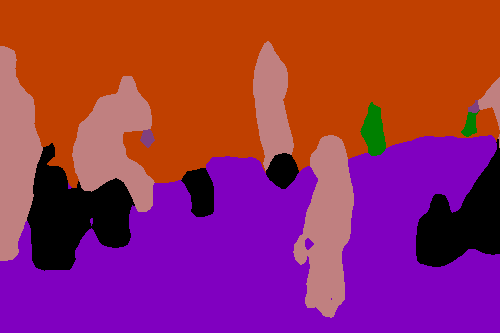} &
\includegraphics[width=\VizImageWidth,height=\VizImageHeightShort]{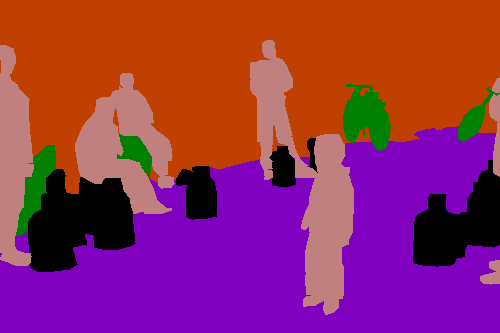} &
\includegraphics[width=\VizImageWidth,height=\VizImageHeightShort]{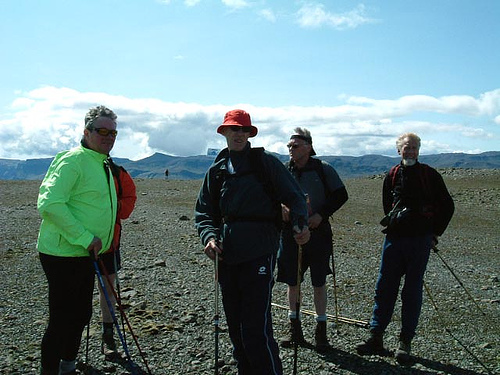} &
\includegraphics[width=\VizImageWidth,height=\VizImageHeightShort]{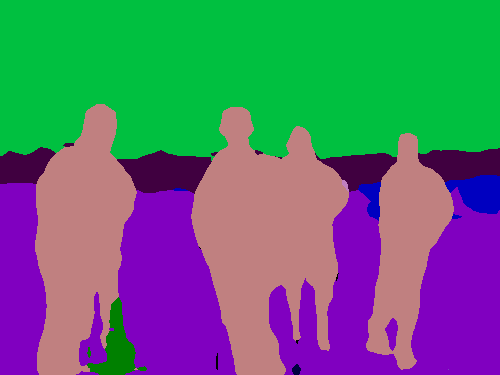} &
\includegraphics[width=\VizImageWidth,height=\VizImageHeightShort]{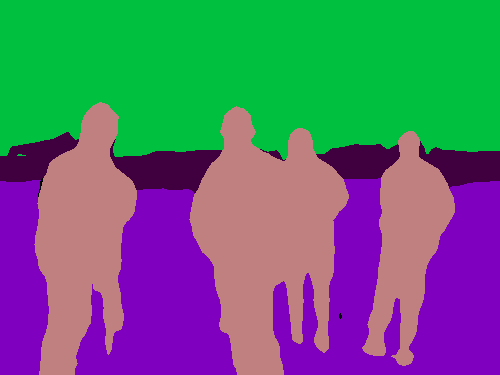} &
\includegraphics[width=\VizImageWidth,height=\VizImageHeightShort]{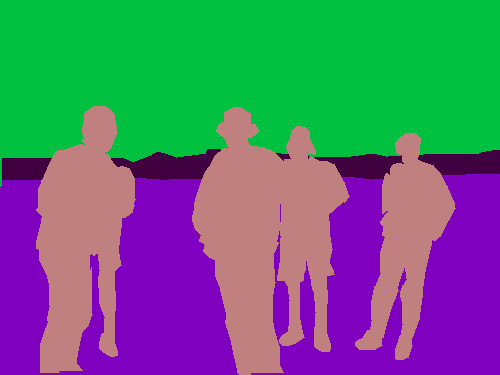} \\
\end{tabular}
\caption{Visualization of the semantic segmentation results on sample PASCAL-Context images. We compare our Holistic FCN$^{+}$ with the FCN method of~\cite{shelhamer_pami16}. We also show the origin images and the ground-truth annotations for reference.}
\label{fig:sup:viz:pascal}
\end{figure*}


\begin{figure*}[tp]
\small
\centering
\tabcolsep 1pt
\begin{tabular}{cccc|cccc}
 & & Holistic & & & & Holistic & \\
Image & DilatedNet~\cite{zhou_arxiv16} & DilatedNet$^{+}$ & Ground Truth & Image & DilatedNet~\cite{zhou_arxiv16} & DilatedNet$^{+}$ & Ground Truth \\
\includegraphics[width=\VizImageWidth,height=\VizImageHeightLong]{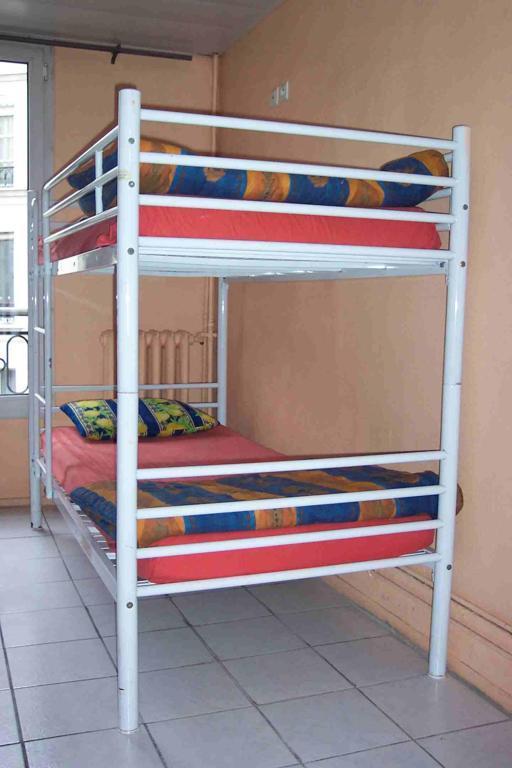} &
\includegraphics[width=\VizImageWidth,height=\VizImageHeightLong]{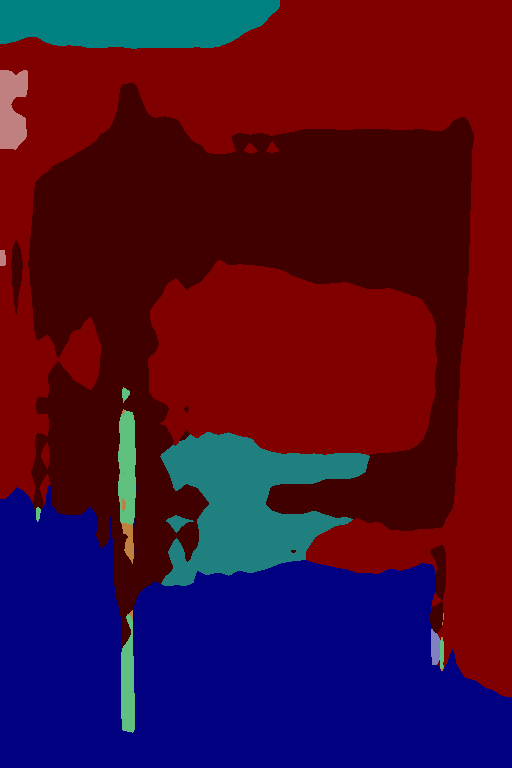} &
\includegraphics[width=\VizImageWidth,height=\VizImageHeightLong]{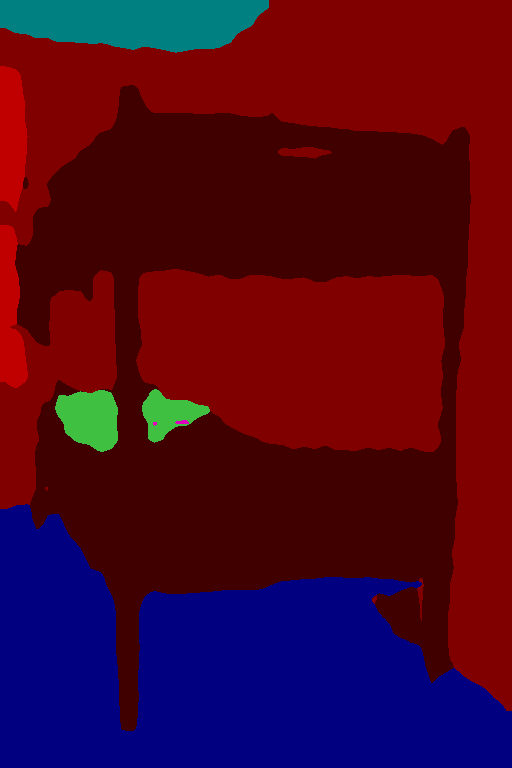} &
\includegraphics[width=\VizImageWidth,height=\VizImageHeightLong]{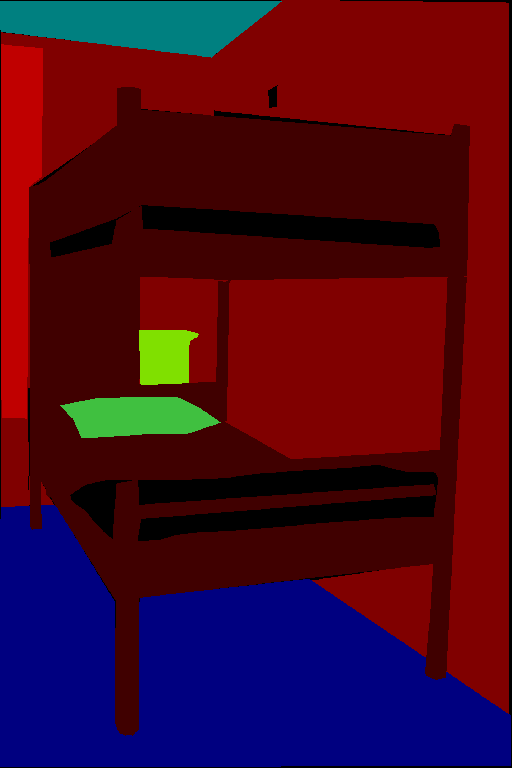} &
\includegraphics[width=\VizImageWidth,height=\VizImageHeightLong]{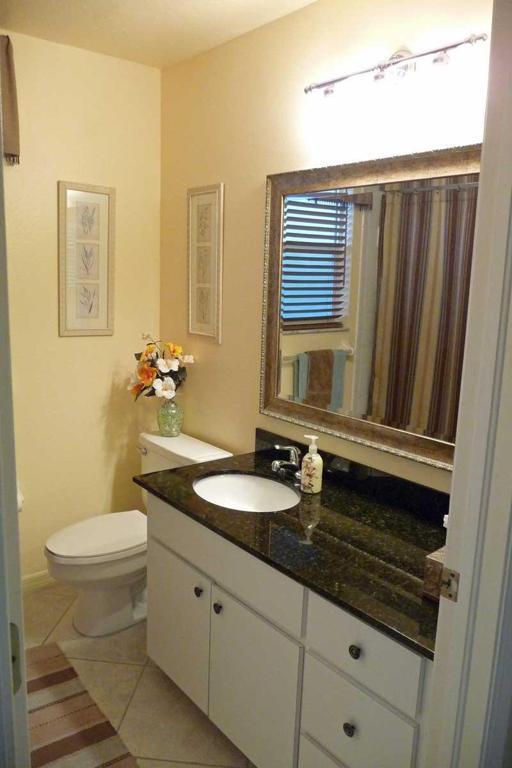} &
\includegraphics[width=\VizImageWidth,height=\VizImageHeightLong]{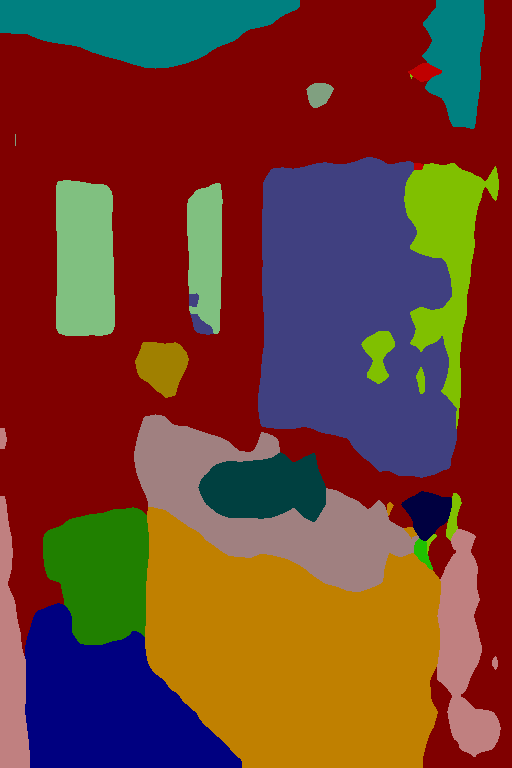} &
\includegraphics[width=\VizImageWidth,height=\VizImageHeightLong]{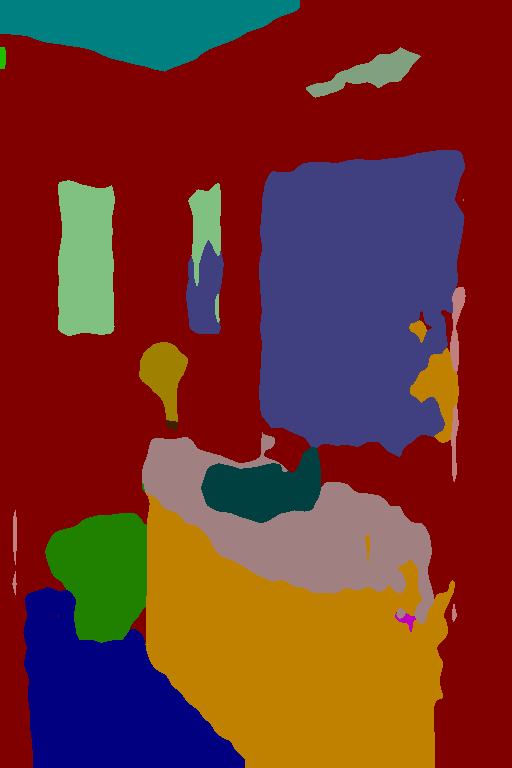} &
\includegraphics[width=\VizImageWidth,height=\VizImageHeightLong]{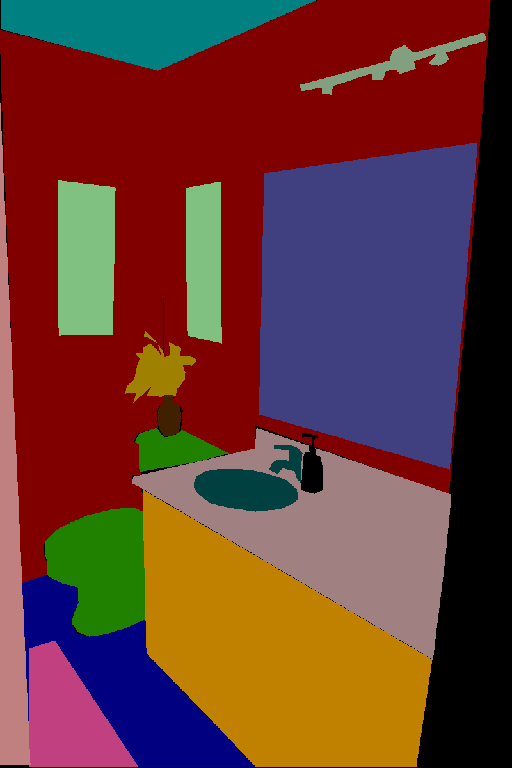} \\
\includegraphics[width=\VizImageWidth,height=\VizImageHeightLong]{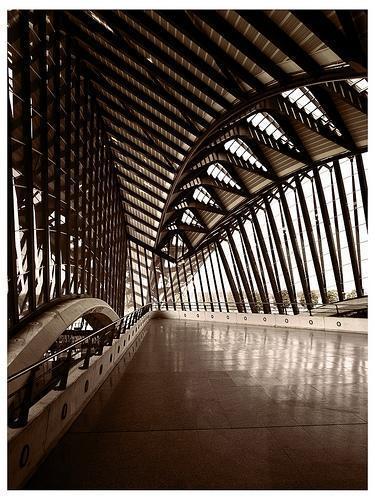} &
\includegraphics[width=\VizImageWidth,height=\VizImageHeightLong]{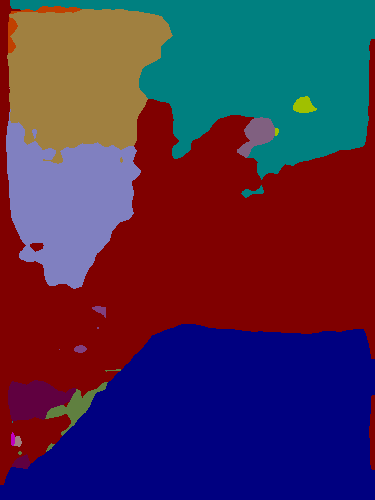} &
\includegraphics[width=\VizImageWidth,height=\VizImageHeightLong]{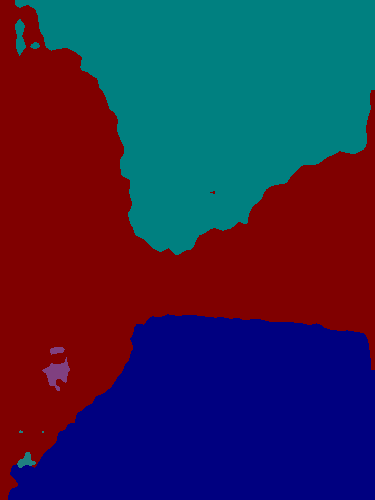} &
\includegraphics[width=\VizImageWidth,height=\VizImageHeightLong]{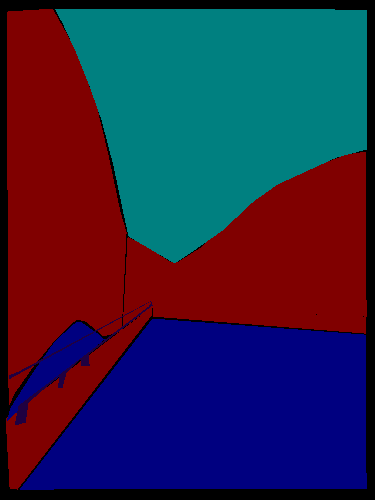} &
\includegraphics[width=\VizImageWidth,height=\VizImageHeightLong]{file/ade/ADE_val_00000028.jpg} &
\includegraphics[width=\VizImageWidth,height=\VizImageHeightLong]{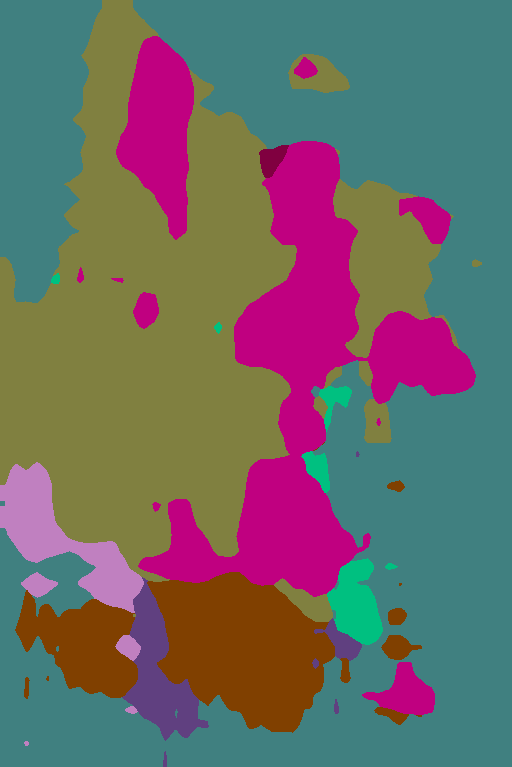} &
\includegraphics[width=\VizImageWidth,height=\VizImageHeightLong]{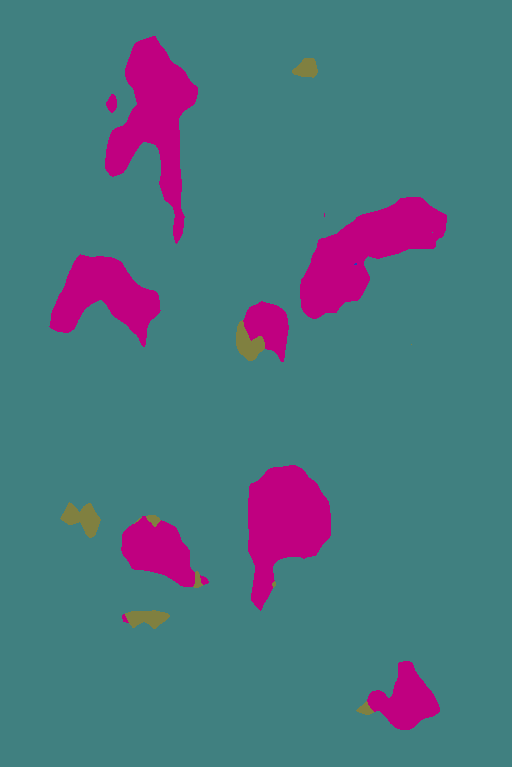} &
\includegraphics[width=\VizImageWidth,height=\VizImageHeightLong]{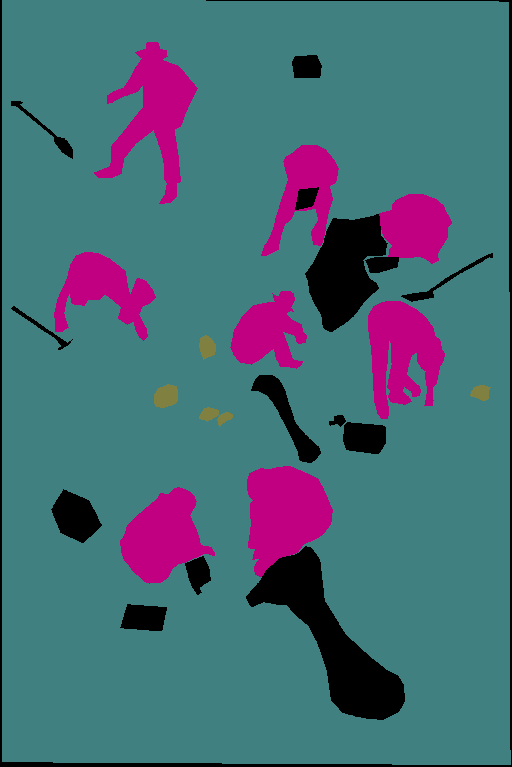} \\
\includegraphics[width=\VizImageWidth,height=\VizImageHeightMiddle]{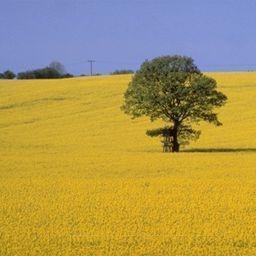} &
\includegraphics[width=\VizImageWidth,height=\VizImageHeightMiddle]{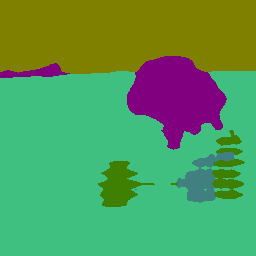} &
\includegraphics[width=\VizImageWidth,height=\VizImageHeightMiddle]{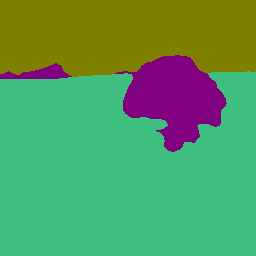} &
\includegraphics[width=\VizImageWidth,height=\VizImageHeightMiddle]{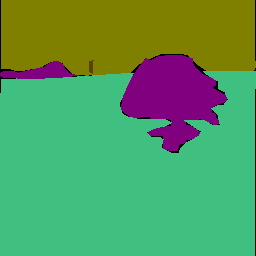} &
\includegraphics[width=\VizImageWidth,height=\VizImageHeightMiddle]{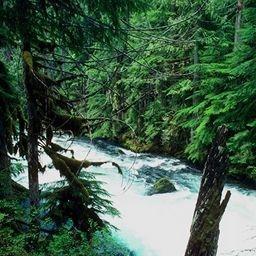} &
\includegraphics[width=\VizImageWidth,height=\VizImageHeightMiddle]{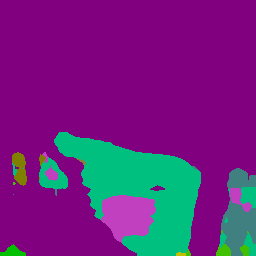} &
\includegraphics[width=\VizImageWidth,height=\VizImageHeightMiddle]{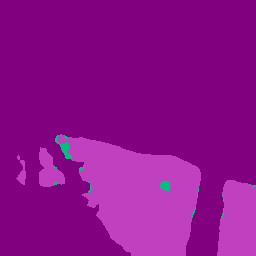} &
\includegraphics[width=\VizImageWidth,height=\VizImageHeightMiddle]{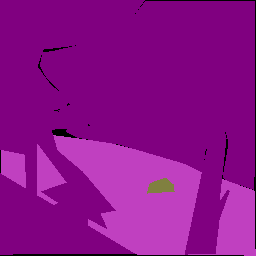} \\
\includegraphics[width=\VizImageWidth,height=\VizImageHeightMiddle]{file/ade/ADE_val_00000767.jpg} &
\includegraphics[width=\VizImageWidth,height=\VizImageHeightMiddle]{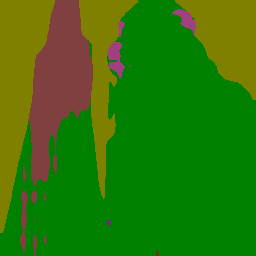} &
\includegraphics[width=\VizImageWidth,height=\VizImageHeightMiddle]{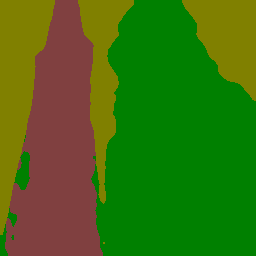} &
\includegraphics[width=\VizImageWidth,height=\VizImageHeightMiddle]{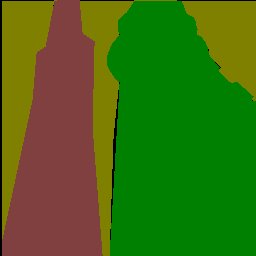} &
\includegraphics[width=\VizImageWidth,height=\VizImageHeightMiddle]{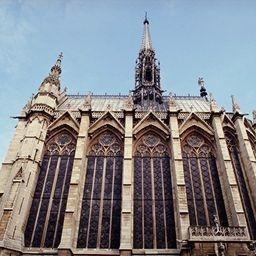} &
\includegraphics[width=\VizImageWidth,height=\VizImageHeightMiddle]{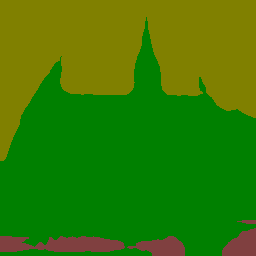} &
\includegraphics[width=\VizImageWidth,height=\VizImageHeightMiddle]{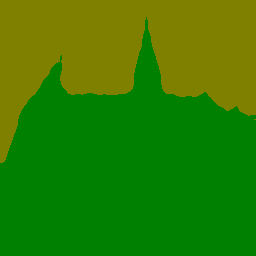} &
\includegraphics[width=\VizImageWidth,height=\VizImageHeightMiddle]{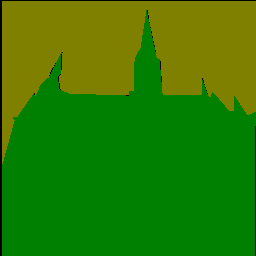} \\
\includegraphics[width=\VizImageWidth,height=\VizImageHeightShort]{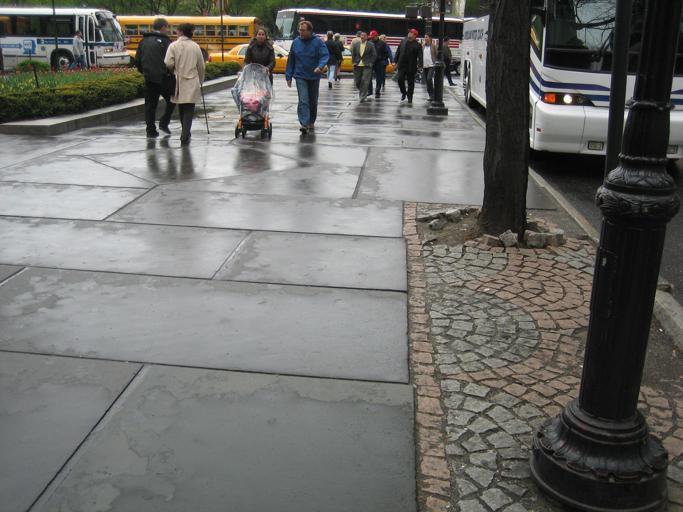} &
\includegraphics[width=\VizImageWidth,height=\VizImageHeightShort]{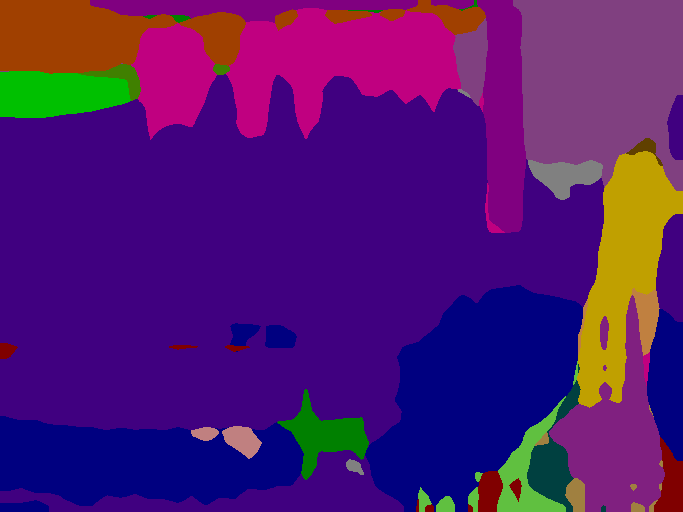} &
\includegraphics[width=\VizImageWidth,height=\VizImageHeightShort]{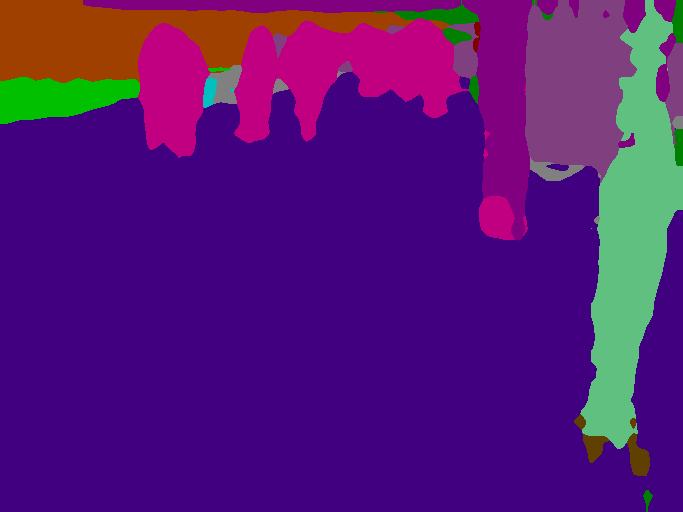} &
\includegraphics[width=\VizImageWidth,height=\VizImageHeightShort]{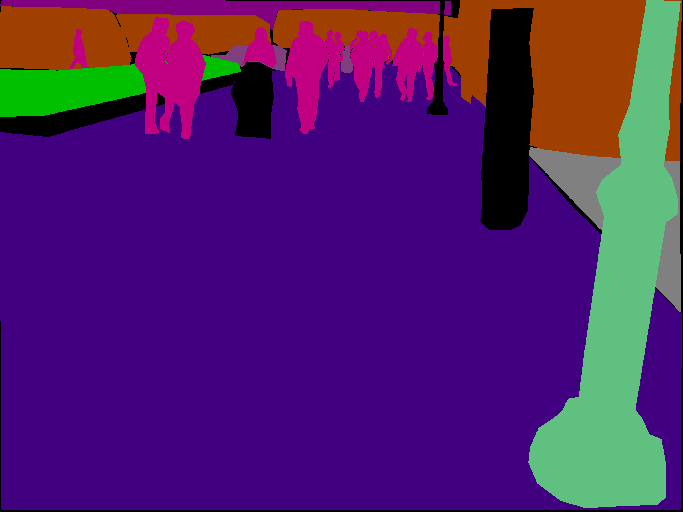} &
\includegraphics[width=\VizImageWidth,height=\VizImageHeightShort]{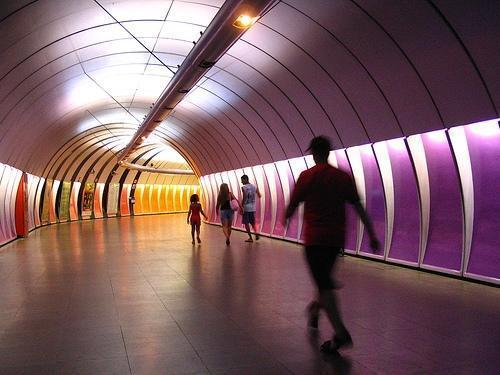} &
\includegraphics[width=\VizImageWidth,height=\VizImageHeightShort]{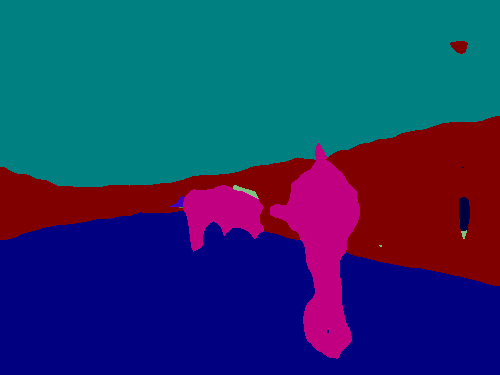} &
\includegraphics[width=\VizImageWidth,height=\VizImageHeightShort]{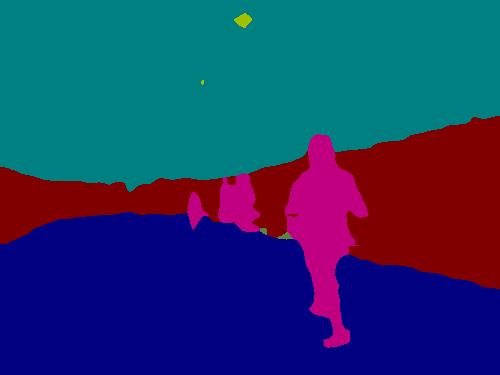} &
\includegraphics[width=\VizImageWidth,height=\VizImageHeightShort]{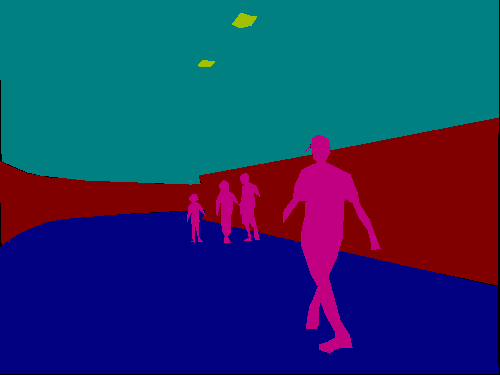} \\
\includegraphics[width=\VizImageWidth,height=\VizImageHeightShort]{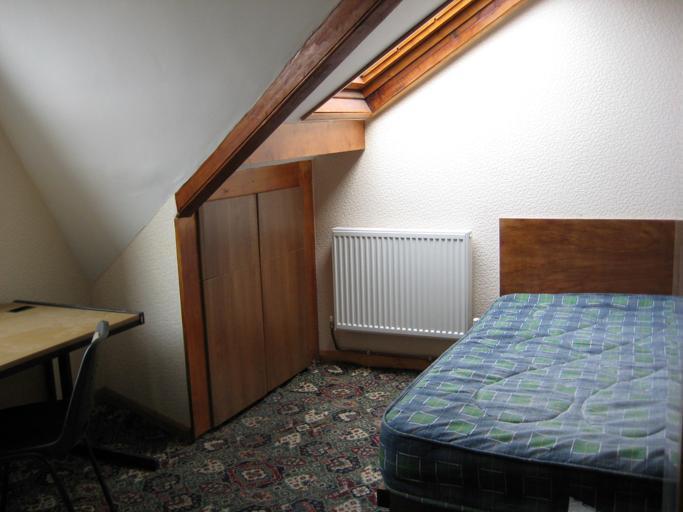} &
\includegraphics[width=\VizImageWidth,height=\VizImageHeightShort]{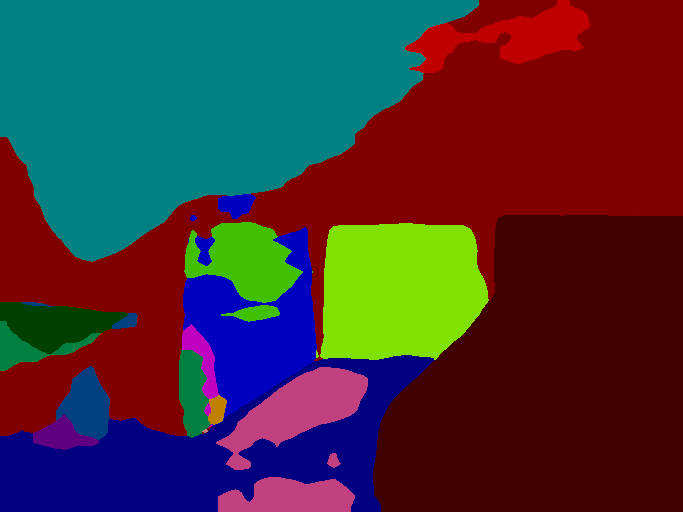} &
\includegraphics[width=\VizImageWidth,height=\VizImageHeightShort]{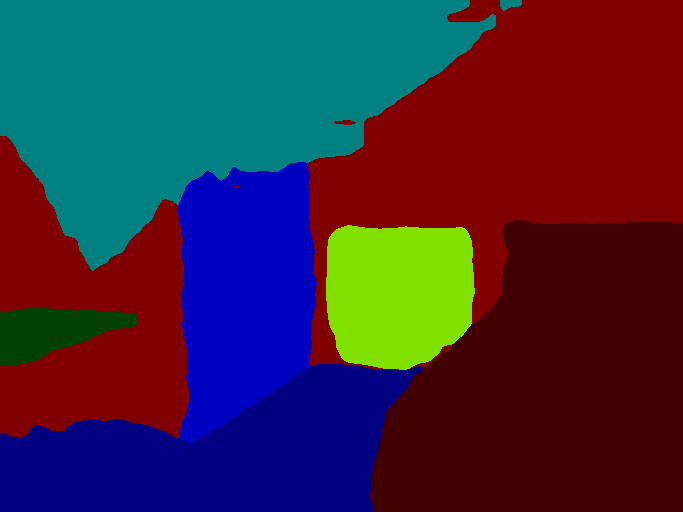} &
\includegraphics[width=\VizImageWidth,height=\VizImageHeightShort]{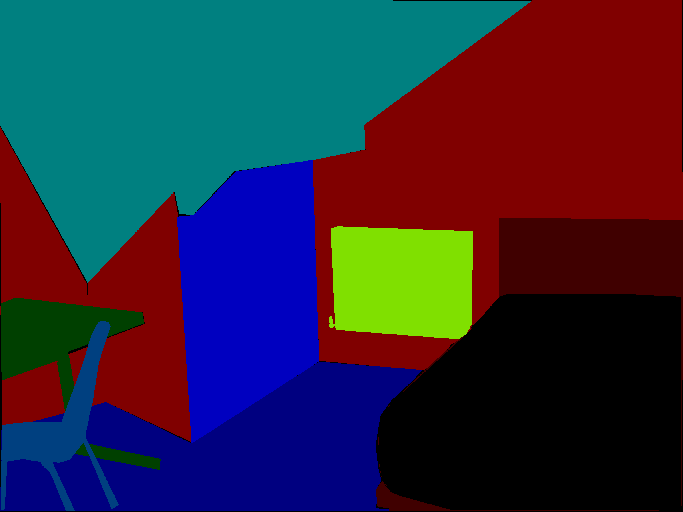} &
\includegraphics[width=\VizImageWidth,height=\VizImageHeightShort]{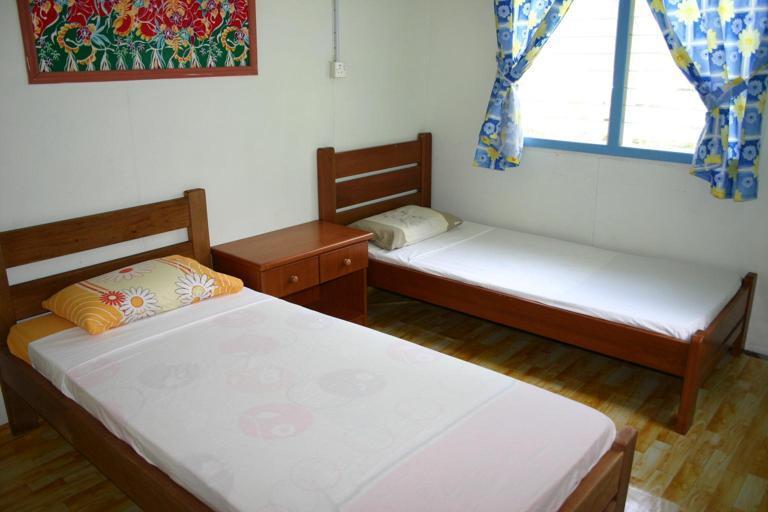} &
\includegraphics[width=\VizImageWidth,height=\VizImageHeightShort]{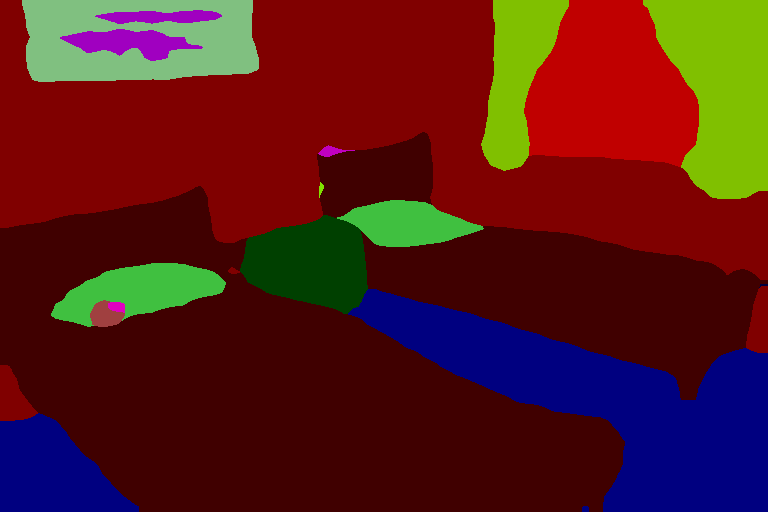} &
\includegraphics[width=\VizImageWidth,height=\VizImageHeightShort]{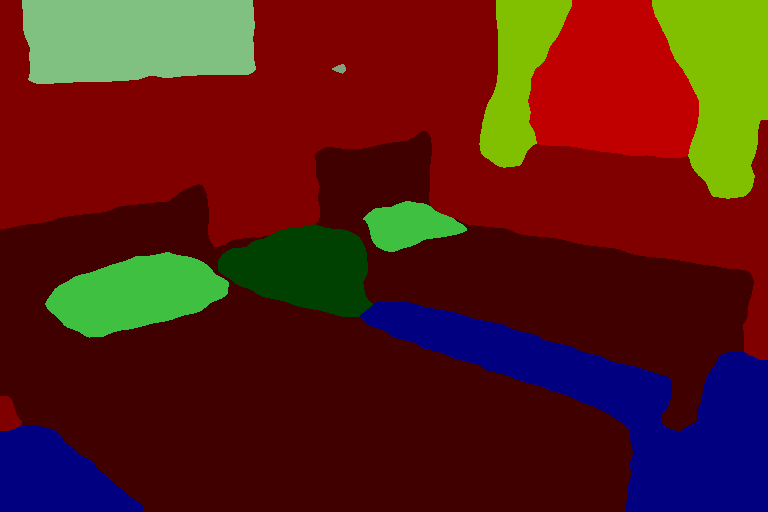} &
\includegraphics[width=\VizImageWidth,height=\VizImageHeightShort]{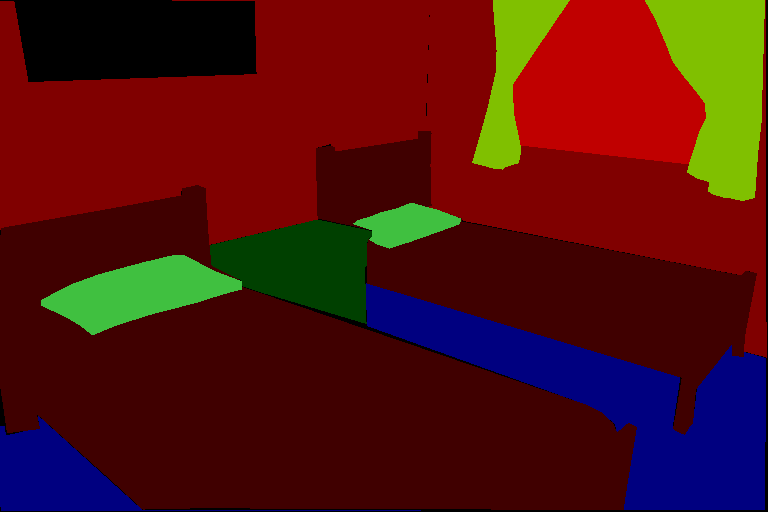} \\
\includegraphics[width=\VizImageWidth,height=\VizImageHeightShort]{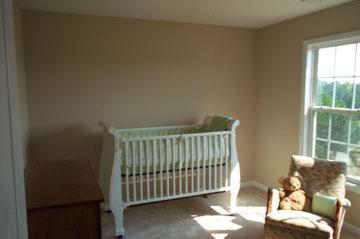} &
\includegraphics[width=\VizImageWidth,height=\VizImageHeightShort]{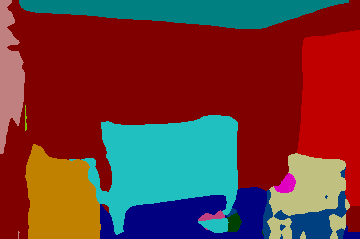} &
\includegraphics[width=\VizImageWidth,height=\VizImageHeightShort]{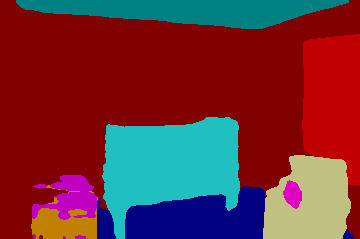} &
\includegraphics[width=\VizImageWidth,height=\VizImageHeightShort]{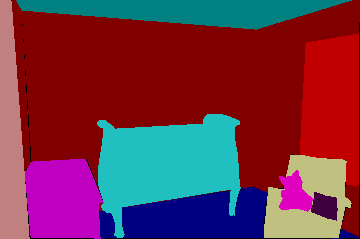} &
\includegraphics[width=\VizImageWidth,height=\VizImageHeightShort]{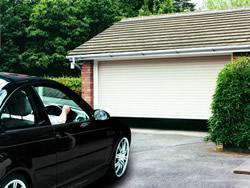} &
\includegraphics[width=\VizImageWidth,height=\VizImageHeightShort]{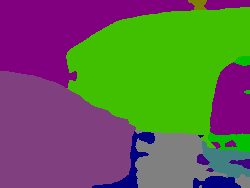} &
\includegraphics[width=\VizImageWidth,height=\VizImageHeightShort]{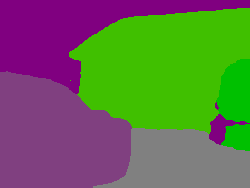} &
\includegraphics[width=\VizImageWidth,height=\VizImageHeightShort]{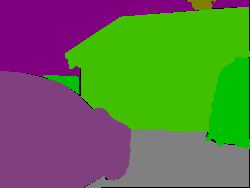} \\
\end{tabular}
\caption{Visualization of the semantic segmentation results on sample ADE20K images. We compare our Holistic DilatedNet$^{+}$ with the DilatedNet method of~\cite{zhou_arxiv16}. We also show the origin images and the ground-truth annotations for reference.}
\label{fig:sup:viz:ade}
\end{figure*}

\end{appendices}

{\small
\bibliographystyle{ieee}
\bibliography{ref}
}

\end{document}